\documentclass[11pt]{article}

\usepackage[final]{acl}

\usepackage{times}
\usepackage{latexsym}

\usepackage[T1]{fontenc}

\usepackage[utf8]{inputenc}

\usepackage{microtype}

\usepackage{inconsolata}

\usepackage{graphicx}
\usepackage{booktabs}
\usepackage{multirow}
\usepackage{makecell}
\usepackage{tabularx}
\usepackage{subcaption}
\usepackage[table]{xcolor}
\usepackage{stfloats}
\usepackage{enumitem}
\usepackage{float}

\usepackage{amsmath,amssymb}

\usepackage{algorithm}
\usepackage{algpseudocode}

\usepackage{xspace}

\newcommand{\algname}{HiVe}
\newcommand{\std}[1]{_{\scriptstyle #1}}

\title{HiVe: Beyond Static Prompts for Multitask Learning \\ via Hierarchy-based Vertical Mixture-of-Experts}

\author{
Hyeonjik Bae \quad Minyeol Kim \quad Susik Yoon \\
Computer Science and Engineering \\
Korea University, Seoul, Korea \\
\texttt{\{bhg4060,minyeol1315,susik\}@korea.ac.kr}
}

\begin{document}
\maketitle
\begin{abstract}
As large language models (LLMs) continue to scale, parameter-efficient fine-tuning (PEFT) has become a practical alternative to full-parameter adaptation. Prompt tuning is effective, but existing approaches either use flat prompt structures or hierarchical structures with fixed prompt composition, limiting adaptive prompt specialization. To address this limitation, we propose \algname{}, a prompt tuning framework that models prompts at multiple levels and enables input-dependent specialization. \algname{} constructs a prompt hierarchy by leveraging inter-task relationships during training, and employs a vertical mixture-of-experts (V-MoE) mechanism at inference time to compose prompts up to the level of specialization required for each input. Experiments show that \algname{} consistently outperforms strong prompt tuning baselines across diverse tasks.
\end{abstract}

\section{Introduction}

The prohibitive cost of full fine-tuning large language models (LLMs) has motivated the development of parameter-efficient fine-tuning (PEFT) that updates a small subset of parameters~\citep{wang-etal-2025-parameter}. Specifically, prompt tuning (PT) learns soft prompts while keeping the backbone model frozen~\citep{lester-etal-2021-power}. The single shared prompt used in early PT approaches has proven ineffective in multi-task and multi-domain scenarios, thereby failing to capture diverse data distributions~\citep{kim-etal-2024-adapromptcl, dun-etal-2025-sweeping}. Recent mixture-of-experts (MoE)-based PT approaches employ input-dependent prompt selection. For example, SMoP~\citep{choi-etal-2023-smop} sparsely routes inputs to short prompts, while PT-MoE~\citep{li-etal-2025-ptmoe} decomposes prompts into low-rank components and combines them through routing. 

\begin{figure}[!t]
\centering
\includegraphics[width=\columnwidth]{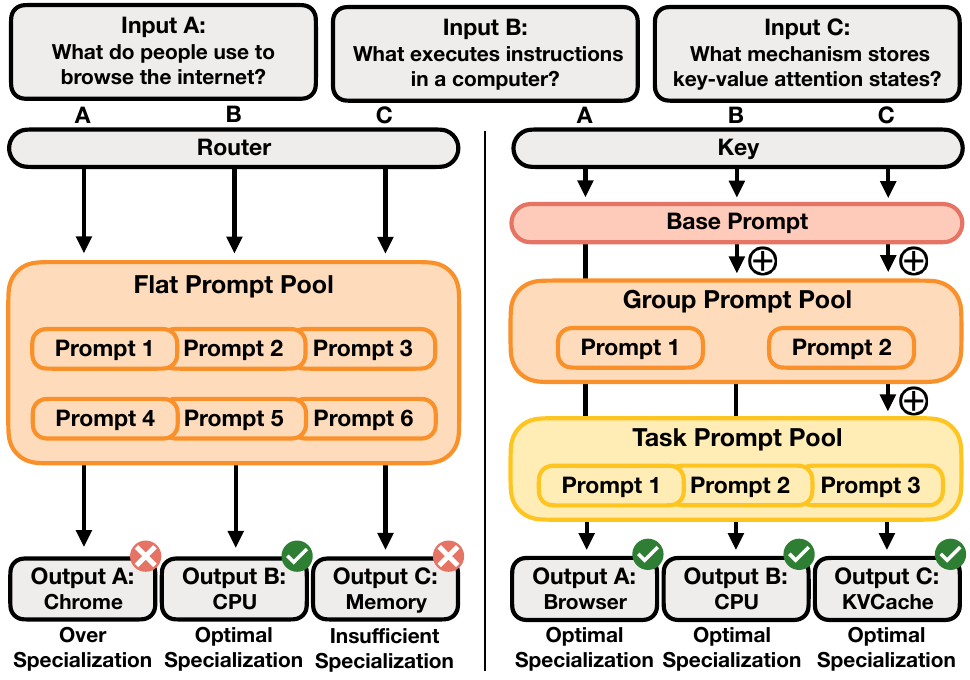}


\makebox[\columnwidth]{
\footnotesize
\hspace{0.3cm}(a) Fixed Specialization.
\hfill
(b) Adaptive Specialization.
}

\caption{
Fixed vs.\ adaptive prompt specialization. Unlike flat MoE approaches (left), \algname{} dynamically constructs and uses the appropriate level of prompts (right).
}

\vspace{-0.2cm}

\label{fig:motivation}
\end{figure}

However, the fundamental limitations of existing approaches lie in their \textit{flat prompt space}, treating all prompts as experts at the same level. As illustrated in Figure~\ref{fig:motivation}, each input requires a diverse set of specializations to produce the correct output; some inputs benefit from general shared knowledge, while others require more task-specific information. Existing flat MoE-based approaches jointly encode both general and specialized knowledge in a few prompts within the same parameter space, making them inadequate for handling varying levels of prompt specialization across inputs.

Prior work has explored multi-level or hierarchical prompting for organizing shared and task-specific knowledge~\citep{wang-etal-2022-hpt, chen-etal-2023-mprompt, liu-etal-2023-hierarchical}, but typically applies a fixed task-dependent path or prompt composition, rather than selecting the hierarchy depth for each input. We identify that the core challenges of realizing adaptive specialization within a hierarchical prompt space are twofold: (i) automatically constructing the prompt hierarchy from learned task relationships at training time, and (ii) dynamically utilizing these disentangled prompts based on input characteristics at inference time. To our knowledge, no existing PT method explicitly addresses both requirements simultaneously.

To this end, we propose \algname{} (Hierarchy-based Vertical Mixture-of-Experts)\footnote{Source code is available at \url{https://github.com/HyeonJikBae/HiVe}.}, a framework that treats a learned prompt hierarchy as a routing space for dynamically selecting the level of specialization required by each input. \algname{} organizes prompts into a hierarchy of progressively specialized representations and employs an input-dependent, key-based prompt routing mechanism.

Specifically, at training time, \algname{} constructs a three-level hierarchy consisting of base, group, and task prompts. However, constructing this hierarchy effectively is challenging, as incorrect grouping can introduce interference between tasks and degrade performance. To mitigate this, we induce a group-level hierarchy based on the stability of learned prompt relationships, allowing semantically related tasks to form stable groups during training. Furthermore, we adopt a residual prompt decomposition strategy that extracts shared information into higher-level prompts while preserving task-specific information at lower levels, thereby reducing both intra- and inter-level redundancy and interference.

At inference time, \algname{} employs a key-based vertical MoE (V-MoE) mechanism that adaptively routes inputs across hierarchical specialization levels. Unlike conventional MoE approaches that route among parallel experts at the same abstraction level, V-MoE determines how deeply to traverse the prompt hierarchy for each input and progressively composes prompts down to the selected level. This enables the model to balance shared and task-specific knowledge according to the specialization required by each input.

We evaluate \algname{} on Machine Reading for Question Answering (MRQA) and summarization benchmarks under both in-domain and out-of-domain settings. Experimental results demonstrate that our hierarchical, adaptive prompt specialization consistently improves both in-domain performance and out-of-domain generalization across diverse tasks over PT baselines. Remarkably, \algname{} with a limited prompt parameter budget performs competitively with resource-heavy alternatives like full fine-tuning ($\sim\!13,000\times$ params) and low-rank adaptation ($\sim\!3.2\times$ params).

In brief, the primary highlights of \algname{} are:
\begin{itemize}[noitemsep, leftmargin=10pt, topsep=2pt]
    \item \textbf{Emergent prompt hierarchy}: A data-driven prompt hierarchy is induced through stability-aware clustering, while residual decomposition separates shared and task-specific representations across hierarchy levels, reducing redundancy and interference.
    \item \textbf{Vertical MoE routing}: A key-based hierarchical routing mechanism that dynamically determines the appropriate level of prompt specialization for each input by composing prompts down to the selected level.
    \item \textbf{Domain-robust generalization}: \algname{} achieves the best results among PT methods, outperforming existing baselines in F1 on 12 MRQA datasets and ROUGE-L on 11 summarization datasets under both in- and out-of-domain settings, while maintaining performance parity with resource-heavy alternatives.
\end{itemize}
\section{Related Work}

\subsection{MoE Prompt Tuning}

MoE has been widely adopted to scale model capacity through sparse expert activation~\citep{shazeer-etal-2017-outrageously}. Recent work has incorporated MoE into parameter-efficient fine-tuning (PEFT) to enable conditional computation over lightweight parameters. AdaMix~\citep{wang-etal-2022-adamix} and HydraLoRA~\citep{tian-etal-2024-hydralora} selectively activate adapter or LoRA experts~\citep{hu-etal-2022-lora}. Prior methods such as ATTEMPT~\citep{asai-etal-2022-attempt}, SMoP~\citep{choi-etal-2023-smop}, and PT-MoE~\citep{li-etal-2025-ptmoe} perform input-dependent prompt composition or routing. However, these approaches rely on flat prompt structures in which all prompts are treated at the same level of specialization, limiting their ability to adaptively model varying levels of specialization across inputs.

\subsection{Task Grouping}

Task relatedness plays a critical role in multi-task learning, particularly in NLP settings~\citep{bingel-etal-2017-identifying}. However, jointly learning loosely related tasks can result in interference and lead to negative transfer~\citep{ni-etal-2023-aggregating}. Prior work has learned task groupings either by representing tasks as sparse combinations of shared latent components~\citep{kumar-etal-2012-learning} or by estimating gradient-based task affinity~\citep{fifty-etal-2021-efficiently}. More recently, task vectors have been used to dynamically group source tasks based on their relevance to a target task in multi-task prompt tuning~\citep{zhang-etal-2025-dynamic}. Building on these approaches, we construct hierarchical prompts based on task relationships learned during training.
\begin{figure*}[!t]
    \centering
    \includegraphics[width=\textwidth]{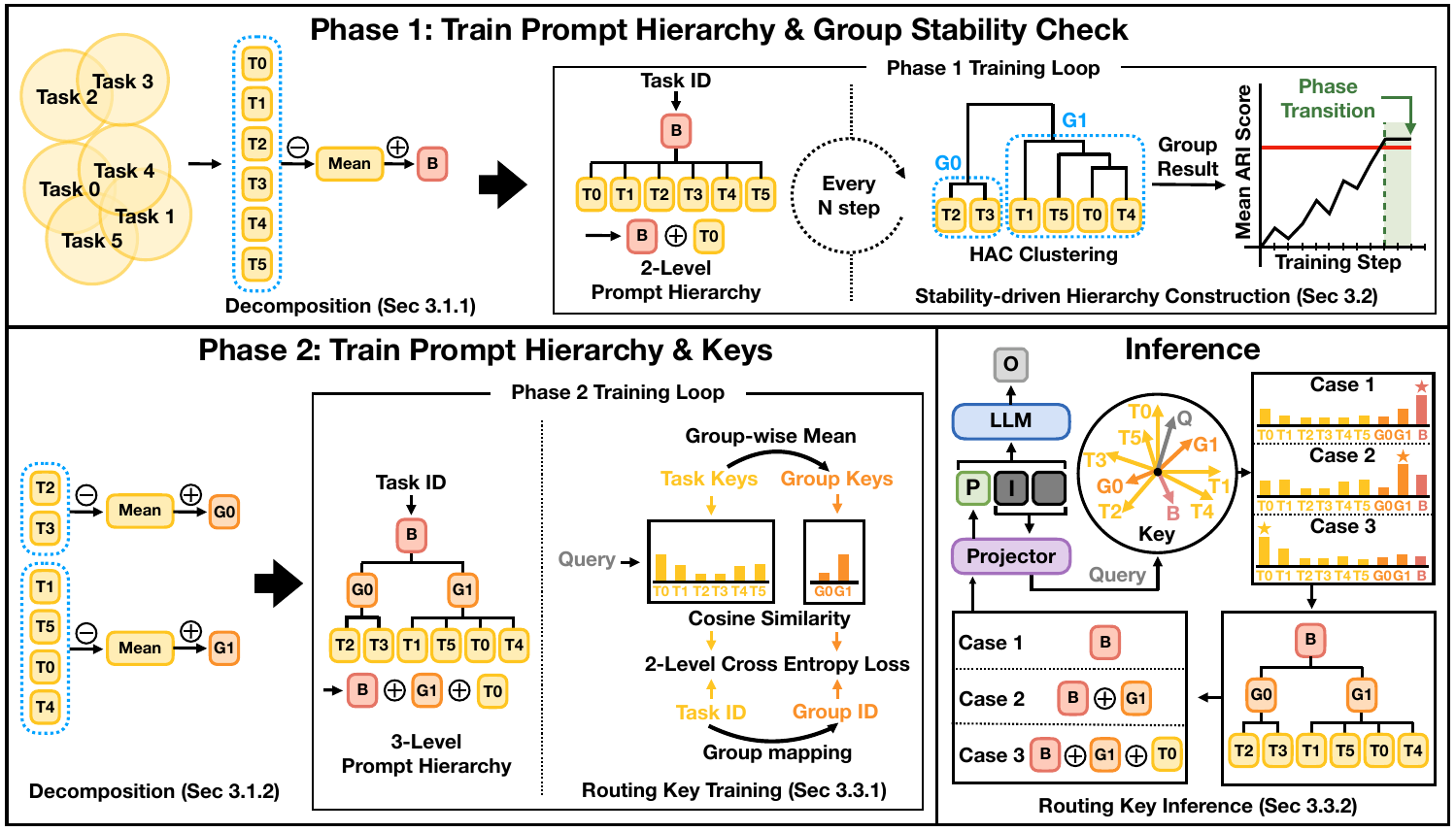}
    \caption{
    Overview of \algname{}, consisting of two training phases followed by inference: (1) stability-driven hierarchy induction, (2) joint optimization of hierarchical prompts and routing keys, and (3) input-dependent hierarchical prompt selection through vertical mixture-of-experts (V-MoE) routing.
    }
    \label{fig:overall_framework}
\end{figure*}

\section{Methods}

Figure~\ref{fig:overall_framework} illustrates the overall framework of \algname{}, rooted in a V-MoE mechanism with a \textit{base-group-task} prompt hierarchy. At inference time, \algname{} matches the input query against routing keys, selects the appropriate specialization level, and constructs the final prompts along the hierarchical path. At training time, after the initialization step described in Appendix~\ref{sec:initial}, \algname{} operates in two training phases. In Phase 1, the model learns a base prompt and task prompts, forming a two-level hierarchy, and clusters tasks based on their task prompts. In Phase 2, group prompts are introduced to complete the three-level hierarchy, and routing keys are jointly optimized to direct each input query to the most appropriate prompt. Full training details are provided in Algorithm~\ref{alg:training}. The following sections introduce the three core components of \algname{}—prompt decomposition, hierarchy construction, and V-MoE—in detail.

\begin{algorithm}[!t]
\scriptsize
\caption{Training of \algname{}}
\label{alg:training}
\begin{algorithmic}[1]

\Require Training data $\mathcal{X}$, template embeddings $T$, task embeddings $E$, low-rank dimension $r$
\Ensure Constructed prompt and key loss

\State $(P_B, P_T, W) \leftarrow
\textsc{InitializePrompt}(T,E,r)$

\State \textcolor{red}{\texttt{/* Sec.~3.1.1 */}}

\State $(P_B, P_T) \leftarrow
\textsc{DecompPrompt}(P_B, P_T)$

\For{each training step}

    \State $\mathcal{L}_{key} \leftarrow 0$ and $(x, t) \sim \mathcal{X}$ \hfill { // $x$: input, $t$: task ID}
    

    \If{Phase 1}

        \State $\tilde{P} \leftarrow
        \textsc{ConstructPrompt}(P_B,P_T,W, t)$

        \State \textcolor{red}{\texttt{/* Sec.~3.2 */}}

        \If{step $\bmod N = 0$}

            \State $c \leftarrow
            \textsc{UpdateGrouping}(P_T)$

            \State $S \leftarrow
            \textsc{ComputeMeanARI}(c)$

            \If{\textsc{IsStableGrouping}$(S)$}

                \State \textcolor{red}{\texttt{/* Sec.~3.1.2 */}}

                \State $(P_G,P_T)\leftarrow
                \textsc{DecompPrompt}(P_T, c)$

                \State $\mathcal{K} \leftarrow \textsc{InitializeKey}(E, W)$

                \State $\texttt{Phase1}
                \rightarrow
                \texttt{Phase2}$

            \EndIf

        \EndIf

    \Else

        \State \textcolor{red}{\texttt{/* Sec.~3.3.1 */}}

        \State $\mathcal{L}_{key} \leftarrow \textsc{TrainKey}(\mathcal{K},t,c)$

        \State $\tilde{P} \leftarrow
        \textsc{ConstructPrompt}(P_B,P_G,P_T,W, t, c)$

    \EndIf

\EndFor
\State \Return
$\tilde{P}, \mathcal{L}_{key}$
\end{algorithmic}
\end{algorithm}

\subsection{Residual Prompt Decomposition}
We apply residual-based prompt decomposition before training to prevent information concentration in lower-level prompts, allowing higher-level prompts to retain general task-solving capability.

\subsubsection{Base-Task Decomposition}

The initial prompt space consists of a base prompt $P_B \in \mathbb{R}^{l \times r}$ and task prompts $P_T = \{P_T^{(\tau)}\}_{\tau=1}^{n}$, where each task prompt $P_T^{(\tau)} \in \mathbb{R}^{l \times r}$ corresponds to an individual task. Here, $n$, $l$, and $r$ denote the number of tasks, prompt length, and low-rank dimension used in the prompt space, respectively.

$P_T^{(\tau)}$ is initialized from task-specific data samples, inherently entangling representations shared across tasks with task-specific representations:
\begin{equation}
    P_T^{(\tau)}
    =
    P_{sh} + P_{sp}^{(\tau)},
\end{equation}
where $P_{sh}$ and $P_{sp}^{(\tau)}$ denote the shared and task-specific representations, respectively.
To extract $P_{sh}$, we estimate it as the global mean of $P_T$:
\begin{equation}
    P_{sh}
    =
    \frac{1}{n}
    \sum_{\tau=1}^{n} P_T^{(\tau)}.
\end{equation}
Then, $P_{sh}$ is transferred from $P_T$ to $P_B$:
\begin{equation}
    P_T^{(\tau)}
    \leftarrow
    P_T^{(\tau)} - P_{sh}, \quad
    P_B
    \leftarrow
    P_B + P_{sh}.
\end{equation}

\subsubsection{Group-Task Decomposition}

Once clustering assignments $c$ are finalized via stability-based clustering (Section~\ref{sec:grouping}), group prompts $P_G = \{P_G^{(g)}\}_{g=1}^{m}$ are constructed based on $c$, where each group prompt $P_G^{(g)} \in \mathbb{R}^{l \times r}$ and $m$ denotes the number of groups.

To extract group-shared representations $P_{sh}^{(g)}$, we estimate them as the mean representation of task prompts in the same group:
\begin{equation}
    P_{sh}^{(g)}
    =
    \frac{1}{|\mathcal{T}_g|}
    \sum_{\tau \in \mathcal{T}_g}
    P_T^{(\tau)},
\end{equation}
where $\mathcal{T}_g = \{\tau \mid c(\tau)=g\}$ denotes the set of task indices assigned to group $g$. Then, $P_{sh}^{(g)}$ is transferred from $P_T$ to $P_G$:
\begin{equation}
    P_T^{(\tau)}
    \leftarrow
    P_T^{(\tau)}
    -
    P_{sh}^{(c(\tau))}, \quad
    P_G^{(g)}
    \leftarrow
    P_{sh}^{(g)}.
\end{equation}

Through these two decompositions, $P_B$, $P_G$, and $P_T$ are encouraged to encode representations at their respective levels of granularity.

\subsection{Stability-driven Hierarchy Construction}
\label{sec:grouping}

During Phase 1, prompts are trained as follows:
\begin{equation}
\tilde{P}_{\mathrm{phase1}}
=
(P_B + P_T^{(t)})W,
\end{equation}
where $t$ denotes the ground-truth task ID and $W \in \mathbb{R}^{r \times d}$ denotes the projection matrix. As training progresses, $P_T$ increasingly encodes task characteristics, serving as the basis for task grouping.

To measure task similarity, the distance matrix $\mathbf{D}$ is computed via cosine dissimilarity between mean-pooled task prompt vectors $\mathbf{v}_i$ and $\mathbf{v}_j$:
\begin{equation}
D_{i,j} = 1 - \frac{\mathbf{v}_i \cdot \mathbf{v}_j}{\|\mathbf{v}_i\| \|\mathbf{v}_j\|}.
\end{equation}
Based on $\mathbf{D}$, tasks are grouped via Hierarchical Agglomerative Clustering (HAC), producing clustering assignment $c_s$ at training step $s$:
\begin{equation}
c_s = \mathrm{HAC}(\mathbf{D}).
\end{equation}
The number of clusters is selected based on the silhouette score over cluster sizes from 2 to n. Singleton clusters are excluded from $P_G$ construction.

Grouping too early may result in inaccurate clustering due to immature prompts, while grouping too late reduces training time for hierarchical learning. To finalize $c$ at the appropriate stage, we evaluate clustering stability via an ARI-based metric~\citep{hubert-etal-1985-comparing}. Every $N$ steps, clustering is performed and $c_s$ is compared against the previous $K$ results to compute $\mathrm{MeanARI}_s$:
\begin{equation}
\mathrm{MeanARI}_s = \frac{1}{K} \sum_{k=1}^{K} \mathrm{ARI}(c_s, c_{s-Nk}).
\end{equation}
Once $\mathrm{MeanARI}_s$ exceeds the threshold $\eta$ for $R$ consecutive checks, $c$ is finalized and Phase 2 begins.

\subsection{Vertical Mixture-of-Experts (V-MoE)}

V-MoE dynamically selects the appropriate level of prompt specialization within the prompt hierarchy for each input. Unlike conventional softmax-based MoE routers, which struggle to model hierarchical relationships, V-MoE employs a key-based routing mechanism that naturally encodes such relationships through centroid derivation.

\subsubsection{Routing Key Training}
In Phase 2, we introduce a routing key set $\mathcal{K} = \{k_B, k_G^{(g)}, k_T^{(\tau)}\}$, where each key lies in the same low-rank space $\mathbb{R}^r$ as the prompts. Task keys $k_T$ are explicitly optimized, and higher-level keys are derived as centroids:
\begin{equation}
k_B = \frac{1}{n} \sum_{\tau=1}^{n} k_T^{(\tau)}, \quad
k_G^{(g)} = \frac{1}{|\mathcal{T}_g|} \sum_{\tau \in \mathcal{T}_g} k_T^{(\tau)}.
\end{equation}
This allows $k_B$ and $k_G$ to capture shared semantics while reducing bias toward individual task representations. The goal of key training is to attract each input query to its task key while repelling others, using derived group keys to preserve intra-group proximity. To this end, the mean-pooled input $x \in \mathbb{R}^d$ is projected via $W$ to obtain query $q = xW^\top \in \mathbb{R}^r$, which is matched against the routing keys via cosine similarity:
\begin{equation}
s_T^{(\tau)} = \cos(q, k_T^{(\tau)}), \quad s_G^{(g)} = \cos(q, k_G^{(g)}).
\end{equation}

The routing keys are optimized via a two-level cross-entropy loss with ground-truth supervision:
\begin{equation}
\mathcal{L}_{key} = \mathrm{CE}(s_T, t) + \mathrm{CE}(s_G, c(t)).
\end{equation}

Independently, the prompt is constructed using the full ground-truth hierarchy:
\begin{equation}
\tilde{P}_{\mathrm{phase2}} = (P_B + P_G^{(c(t))} + P_T^{(t)})W.
\end{equation}
This decoupled training ensures that all hierarchy levels are jointly utilized during training, allowing each level to learn appropriate representations while preventing potential knowledge contamination from routing errors.

\subsubsection{Routing Key Inference}
Without requiring task ID, the model selects the routing key $k^*$ with the highest cosine similarity to $q$ among all keys in $\mathcal{K}$:
\begin{equation}
k^*
=
\arg\max_{k \in \mathcal{K}}
\cos(q, k),
\end{equation}
where $k^*$ determines the hierarchical level, and the prompt is composed by aggregating prompts down to the selected level, ranging from $P_B$ alone to the full prompt hierarchy $P_B + P_G^{(g)} + P_T^{(\tau)}$:
\begin{equation}
\hspace{-0.1em}
\tilde{P}_{\mathrm{inf}} = \left\{
\begin{aligned}
&P_B,
&& \mkern-10mu k^* = k_B \\
&P_B + P_G^{(g^*)},
&& \mkern-10mu k^* = k_G^{(g^*)} \\
&P_B + P_G^{(c(\tau^*))}\!+\!P_T^{(\tau^*)}\mkern-5mu.
&& \mkern-10mu k^* = k_T^{(\tau^*)}
\end{aligned}
\right.
\end{equation}
After constructing a prompt at an appropriate specialization level, $\tilde{P}_{\mathrm{inf}}$ is projected through $W$ to obtain the final prompt representation $\tilde{P}_{\mathrm{inf}}W$, which is prepended to the input sequence.

\section{Experiments}

\subsection{Experimental Setup}

\textbf{Datasets and Tasks.} \algname{} is evaluated across question answering and summarization tasks. For question answering, we use the MRQA benchmark~\citep{fisch-etal-2019-mrqa}, consisting of 6 in-domain and 6 out-of-domain datasets. For summarization, we construct a diverse benchmark composed of 6 in-domain and 5 out-of-domain datasets. For in-domain settings, models are trained on the training splits and periodically evaluated on the validation splits every few training steps. The checkpoint with the best validation performance is selected and evaluated directly on the out-of-domain datasets without further training. Detailed dataset descriptions are provided in Appendix~\ref{sec:dataset_details}.

\textbf{Baselines and Model.} We compare \algname{} with representative adaptation methods, including full fine-tuning (FT), parameter-efficient fine-tuning (PEFT) methods such as LoRA~\citep{hu-etal-2022-lora} and HydraLoRA~\citep{tian-etal-2024-hydralora}, and prompt-based methods such as Prompt Tuning (PT)~\citep{lester-etal-2021-power}, DPT~\citep{xiao-etal-2023-decomposed}, SMoP~\citep{choi-etal-2023-smop}, and PT-MoE~\citep{li-etal-2025-ptmoe}. An adapted version of HiPro~\citep{liu-etal-2023-hierarchical} is evaluated only in-domain, as it requires task IDs at inference. Baselines are selected to support both in- and out-of-domain evaluation under a unified input and backbone architecture. All methods use Llama-3.2-1B-Instruct as the backbone. For fair comparison, all prompt-based methods are configured to use a comparable number of trainable parameters (approximately 87K), while LoRA-based baselines are adjusted to use similar budgets whenever possible. Detailed hyperparameter settings are provided in Appendix~\ref{sec:baseline_hyperparameters}.

\textbf{Evaluation Metrics.} Evaluation follows the standard protocols for each benchmark. For question answering, we report F1 scores in the main results and provide Exact Match (EM) scores in Appendix~\ref{sec:mrqa_em_results}. For summarization, we report ROUGE-L scores. All results are averaged over three runs with different random seeds.

\begin{table*}[!t]
\centering
\scriptsize
\setlength{\tabcolsep}{1.2pt}
\renewcommand{\arraystretch}{1.0}
\vspace{-0.5em}

\begin{tabular}{
l c
c c c c c c >{\columncolor{gray!8}}c
c c c c c c >{\columncolor{gray!8}}c
}
\toprule
\multirow{2}{*}{\textbf{Method}}
& \multirow{2}{*}{\makecell{\textbf{\# of} \\ \textbf{Params}}}
& \multicolumn{7}{c}{\textbf{In-domain}}
& \multicolumn{7}{c}{\textbf{Out-of-domain}} \\
\cmidrule(lr){3-9}
\cmidrule(lr){10-16}
&
& SQ & News & Tri & Srch & HP & NQ & Avg
& BSQ & DP & DRC & RC & RE & TB & Avg \\
\midrule

FT
& 1.2B
& $87.3\std{0.9}$
& $60.4\std{0.7}$
& $71.7\std{1.1}$
& $81.8\std{0.5}$
& $75.7\std{0.9}$
& $74.0\std{0.6}$
& $75.2\std{0.2}$
& $73.5\std{2.4}$
& $48.6\std{0.8}$
& $49.0\std{1.4}$
& $47.0\std{0.8}$
& $85.6\std{0.8}$
& $56.8\std{2.6}$
& $60.1\std{1.1}$
\\

\midrule

LoRA
& 106K
& $87.3\std{0.4}$
& $59.2\std{0.6}$
& $74.8\std{0.3}$
& $79.4\std{1.0}$
& $74.4\std{0.5}$
& $73.9\std{0.3}$
& $74.9\std{0.3}$
& $77.0\std{0.7}$
& $47.5\std{1.5}$
& $49.8\std{0.3}$
& $50.4\std{0.4}$
& $85.3\std{0.5}$
& $60.0\std{0.9}$
& $61.7\std{0.7}$
\\

HydraLoRA
& 278K
& $87.8\std{0.6}$
& $59.0\std{0.5}$
& $74.9\std{0.4}$
& $79.8\std{1.5}$
& $74.9\std{0.4}$
& $74.3\std{0.2}$
& $75.2\std{0.4}$
& $76.6\std{0.5}$
& $48.6\std{2.6}$
& $49.7\std{0.3}$
& $51.0\std{1.3}$
& $85.3\std{0.7}$
& $58.8\std{1.4}$
& $61.7\std{0.3}$
\\

\midrule

HiPro-adapted
& 75K / 90K
& $88.8\std{0.2}$
& $57.4\std{0.3}$
& $72.5\std{0.2}$
& $77.6\std{0.2}$
& $75.4\std{0.3}$
& $74.6\std{0.4}$
& $74.4\std{0.2}$
& --
& --
& --
& --
& --
& --
& --
\\

\midrule

PT
& 81K
& $87.1\std{0.6}$
& $56.5\std{1.7}$
& $71.9\std{3.9}$
& $73.5\std{1.4}$
& $72.5\std{0.7}$
& $73.1\std{1.1}$
& $72.5\std{1.0}$
& $76.6\std{0.4}$
& $48.9\std{0.6}$
& $46.9\std{0.3}$
& $\mathbf{50.4}\std{0.2}$
& $\underline{86.0}\std{0.3}$
& $56.2\std{2.3}$
& $60.8\std{0.2}$
\\

DPT
& 87K
& $86.8\std{0.3}$
& $55.8\std{3.7}$
& $\underline{72.5}\std{1.7}$
& $77.2\std{2.2}$
& $73.1\std{0.8}$
& $73.3\std{0.9}$
& $73.1\std{1.5}$
& $76.8\std{0.5}$
& $\underline{51.3}\std{0.6}$
& $48.6\std{0.1}$
& $50.2\std{0.2}$
& $85.5\std{0.6}$
& $\underline{58.5}\std{0.9}$
& $\underline{61.8}\std{0.2}$
\\

SMoP
& 86K
& $\underline{87.5}\std{0.7}$
& $55.4\std{3.0}$
& $72.0\std{1.1}$
& $\underline{78.3}\std{1.1}$
& $73.1\std{0.6}$
& $\underline{73.5}\std{0.8}$
& $\underline{73.3}\std{0.6}$
& $\underline{77.1}\std{1.3}$
& $51.0\std{1.1}$
& $47.8\std{1.4}$
& $50.1\std{1.9}$
& $85.9\std{0.4}$
& $56.5\std{2.3}$
& $61.4\std{0.7}$
\\

PT-MoE
& 87K
& $87.0\std{0.7}$
& $\underline{57.4}\std{0.7}$
& $72.0\std{1.7}$
& $75.2\std{1.4}$
& $\underline{73.2}\std{0.5}$
& $72.1\std{1.3}$
& $72.8\std{0.2}$
& $76.2\std{0.6}$
& $45.7\std{0.9}$
& $\mathbf{49.2}\std{1.0}$
& $\underline{50.2}\std{0.9}$
& $85.5\std{0.5}$
& $56.6\std{0.7}$
& $60.6\std{0.4}$
\\

\rowcolor{gray!12}
\textbf{\algname{}}
& 87K
& $\mathbf{87.7}\std{0.8}$
& $\mathbf{59.2}\std{0.9}$
& $\mathbf{74.9}\std{1.7}$
& $\mathbf{80.5}\std{0.5}$
& $\mathbf{75.2}\std{0.6}$
& $\mathbf{74.9}\std{0.5}$
& $\mathbf{75.4}\std{0.3}$
& $\mathbf{78.3}\std{0.9}$
& $\mathbf{53.9}\std{0.2}$
& $\underline{48.9}\std{0.4}$
& $50.1\std{0.8}$
& $\mathbf{86.0}\std{1.0}$
& $\mathbf{59.8}\std{0.9}$
& $\mathbf{62.8}\std{0.3}$
\\

\bottomrule
\end{tabular}

\vspace{-0.5em}
\caption{
MRQA F1 comparison on in- and out-of-domain datasets with mean and standard deviation.
}
\label{tab:f1_results_std}

\end{table*}

\begin{table*}[!t]
\centering
\scriptsize
\setlength{\tabcolsep}{2.0pt}
\renewcommand{\arraystretch}{1.0}
\vspace{-0.5em}

\begin{tabular}{
l c
c c c c c c >{\columncolor{gray!8}}c
c c c c c >{\columncolor{gray!8}}c
}
\toprule
\multirow{2}{*}{\textbf{Method}}
& \multirow{2}{*}{\makecell{\textbf{\# of} \\ \textbf{Params}}}
& \multicolumn{7}{c}{\textbf{In-domain}}
& \multicolumn{6}{c}{\textbf{Out-of-domain}} \\
\cmidrule(lr){3-9}
\cmidrule(lr){10-15}
&
& Wiki & Red & XS & CNN & Giga & Media & Avg
& Dialog & Sci & AESLC & WH & MN & Avg \\
\midrule

FT
& 1.2B
& $32.0\std{0.7}$
& $20.0\std{0.3}$
& $26.6\std{0.4}$
& $23.0\std{1.5}$
& $39.4\std{0.5}$
& $14.9\std{1.1}$
& $26.0\std{0.6}$
& $22.0\std{1.6}$
& $18.5\std{2.0}$
& $14.8\std{0.4}$
& $22.0\std{1.6}$
& $10.7\std{1.2}$
& $17.6\std{0.5}$
\\

\midrule

LoRA
& 106K
& $30.2\std{0.5}$
& $20.4\std{0.4}$
& $25.3\std{0.5}$
& $23.9\std{0.4}$
& $38.1\std{1.1}$
& $14.0\std{0.5}$
& $25.3\std{0.4}$
& $14.8\std{1.3}$
& $19.0\std{1.3}$
& $14.6\std{1.3}$
& $16.9\std{3.0}$
& $10.6\std{0.5}$
& $15.2\std{0.4}$
\\

HydraLoRA
& 278K
& $30.4\std{0.4}$
& $20.3\std{0.3}$
& $25.7\std{0.4}$
& $23.9\std{0.4}$
& $38.6\std{0.9}$
& $14.3\std{0.6}$
& $25.5\std{0.4}$
& $15.9\std{0.7}$
& $18.1\std{0.4}$
& $14.6\std{0.6}$
& $20.2\std{0.5}$
& $10.2\std{0.4}$
& $15.8\std{0.4}$
\\

\midrule

HiPro-adapted
& 75K / 90K
& $29.5\std{0.3}$
& $20.6\std{0.3}$
& $25.2\std{0.3}$
& $25.0\std{0.2}$
& $38.5\std{0.3}$
& $16.0\std{0.6}$
& $25.8\std{0.3}$
& --
& --
& --
& --
& --
& --
\\

\midrule

PT
& 81K
& $30.0\std{0.4}$
& $20.2\std{0.3}$
& $25.0\std{0.4}$
& $21.6\std{0.6}$
& $38.0\std{0.4}$
& $\underline{16.9}\std{0.4}$
& $25.3\std{0.4}$
& $\mathbf{18.8}\std{1.5}$
& $18.6\std{1.1}$
& $13.1\std{0.8}$
& $19.9\std{0.4}$
& $8.8\std{0.5}$
& $15.8\std{0.4}$
\\

DPT
& 87K
& $30.3\std{0.3}$
& $20.0\std{0.4}$
& $\underline{25.3}\std{0.3}$
& $21.9\std{0.4}$
& $38.0\std{0.4}$
& $16.9\std{0.4}$
& $\underline{25.4}\std{0.3}$
& $18.6\std{0.8}$
& $18.1\std{0.9}$
& $14.1\std{0.8}$
& $\underline{20.8}\std{0.6}$
& $8.9\std{0.4}$
& $16.1\std{0.4}$
\\

SMoP
& 86K
& $\underline{30.4}\std{0.4}$
& $19.8\std{0.6}$
& $24.7\std{0.3}$
& $\underline{23.2}\std{0.5}$
& $37.9\std{0.4}$
& $15.8\std{0.4}$
& $25.3\std{0.3}$
& $18.1\std{0.6}$
& $\mathbf{20.2}\std{1.1}$
& $14.1\std{0.4}$
& $19.0\std{0.5}$
& $\underline{9.1}\std{0.4}$
& $16.1\std{0.3}$
\\

PT-MoE
& 87K
& $29.2\std{0.4}$
& $\underline{20.5}\std{0.4}$
& $25.0\std{0.4}$
& $20.3\std{0.6}$
& $\underline{38.4}\std{0.4}$
& $\mathbf{17.0}\std{0.4}$
& $25.1\std{0.5}$
& $\underline{18.8}\std{0.4}$
& $18.3\std{0.6}$
& $\underline{14.3}\std{1.0}$
& $17.9\std{1.3}$
& $8.1\std{0.6}$
& $15.5\std{0.7}$
\\

\rowcolor{gray!12}
\textbf{\algname{}}
& 87K
& $\mathbf{30.5}\std{0.3}$
& $\mathbf{20.6}\std{0.7}$
& $\mathbf{25.9}\std{0.2}$
& $\mathbf{24.0}\std{0.5}$
& $\mathbf{38.8}\std{0.1}$
& $16.7\std{2.5}$
& $\mathbf{26.1}\std{0.6}$
& $17.5\std{1.6}$
& $\underline{18.7}\std{2.0}$
& $\mathbf{14.4}\std{0.1}$
& $\mathbf{20.9}\std{1.0}$
& $\mathbf{12.3}\std{0.6}$
& $\mathbf{16.8}\std{0.3}$
\\

\bottomrule
\end{tabular}

\vspace{-0.5em}
\caption{
Summarization ROUGE-L comparison on in- and out-of-domain datasets with mean and standard deviation.
}
\label{tab:rouge_results_std}

\end{table*}

\subsection{Experimental Results}

\subsubsection{Overall Performance}

\textbf{MRQA.}
Table~\ref{tab:f1_results_std} presents the main results on the MRQA benchmark. \algname{} achieved the best performance among prompt-based methods in both in-domain and out-of-domain (OOD) settings. In the in-domain setting, it achieved an average F1 score of 75.4, compared with FT (75.2) while updating only approximately 0.007\% of the model parameters. It also achieved the best performance across all in-domain tasks, improving the average F1 score over SMoP and PT-MoE by 2.1 and 2.6 points, respectively. Notably, \algname{} also outperforms the hierarchical baseline HiPro-adapted by 1.0 F1 points. These gains persisted in OOD settings, indicating improved robustness under domain shifts. \algname{} outperformed FT by 2.7 F1 points on average and achieved the best performance on four out of six OOD datasets. In addition, it improved the average F1 score over SMoP and PT-MoE by 1.4 and 2.2 points, respectively. These results suggest that \algname{} effectively adapts prompt specialization levels across tasks and domains.

\textbf{Summarization.}
Table~\ref{tab:rouge_results_std} presents the ROUGE-L results on the summarization benchmark. Similar trends were observed on summarization tasks. In the in-domain setting, \algname{} achieved an average ROUGE-L score of 26.1, compared with FT (26.0). It also achieved the best performance on five out of six in-domain tasks, improving the average ROUGE-L score over SMoP and PT-MoE by 0.8 and 1.0 points, respectively. Relative to HiPro-adapted, \algname{} yields a 0.3-point higher average ROUGE-L score. These results indicate that the improvements are maintained across diverse in-domain summarization datasets. In OOD settings, \algname{} improved the average ROUGE-L score over SMoP and PT-MoE by 0.7 and 1.3 points, respectively, with large gains observed on the Multi-News dataset, the most challenging task in our benchmark. The OOD results further demonstrate stable performance under distribution shifts. These results suggest that \algname{} performs effectively across both extractive and generative tasks.

\begin{table}[!t]
\centering
\footnotesize
\setlength{\tabcolsep}{2.5pt}
\renewcommand{\arraystretch}{0.1}
\newcommand{\down}[1]{\textcolor{blue}{{\scriptsize (#1)}}}
\newcommand{\up}[1]{\textcolor{red}{{\scriptsize (+#1)}}}
\newcommand{\same}{{\scriptsize (0.0)}}
\begin{tabular}{lcccc}
\toprule
\multirow{2}{*}{\textbf{Model}}
& \multicolumn{2}{c}{\textbf{MRQA}}
& \multicolumn{2}{c}{\textbf{Summ.}} \\
\cmidrule(lr){2-3}
\cmidrule(lr){4-5}
& \textbf{In} & \textbf{Out} & \textbf{In} & \textbf{Out} \\
\midrule
\rowcolor{gray!10}
\textbf{\algname{}}
& \textbf{75.4}$\std{0.3}$
& \textbf{62.8}$\std{0.3}$
& \textbf{26.1}$\std{0.6}$
& \textbf{16.8}$\std{0.3}$ \\
\midrule
\textit{w/o Residual Decomp.}
& \makecell[c]{74.9$\std{0.5}$ \\ \down{-0.5}}
& \makecell[c]{61.8$\std{0.2}$ \\ \down{-1.0}}
& \makecell[c]{26.0$\std{0.3}$ \\ \down{-0.1}}
& \makecell[c]{14.4$\std{0.4}$ \\ \down{-2.4}} \\
\midrule
\multicolumn{5}{l}{\textit{w/o Hierarchy}} \\
\quad Base-level only
& \makecell[c]{73.6$\std{0.7}$ \\ \down{-1.8}}
& \makecell[c]{62.1$\std{0.2}$ \\ \down{-0.7}}
& \makecell[c]{25.4$\std{0.1}$ \\ \down{-0.7}}
& \makecell[c]{16.0$\std{0.4}$ \\ \down{-0.8}} \\
\quad Group-level only
& \makecell[c]{73.7$\std{0.7}$ \\ \down{-1.7}}
& \makecell[c]{61.1$\std{0.8}$ \\ \down{-1.7}}
& \makecell[c]{25.5$\std{0.1}$ \\ \down{-0.6}}
& \makecell[c]{15.2$\std{1.0}$ \\ \down{-1.6}} \\
\quad Task-level only
& \makecell[c]{74.4$\std{0.6}$ \\ \down{-1.0}}
& \makecell[c]{61.6$\std{0.4}$ \\ \down{-1.2}}
& \makecell[c]{26.1$\std{0.2}$ \\ \same}
& \makecell[c]{14.5$\std{0.6}$ \\ \down{-2.3}} \\
\midrule
\multicolumn{5}{l}{\textit{w/o V-MoE}} \\ \\[5pt]
\quad \makecell[l]{Base-key only \\ {\scriptsize (Always B)}}
& \makecell[c]{71.1$\std{0.1}$ \\ \down{-4.3}}
& \makecell[c]{63.1$\std{0.2}$ \\ \up{0.3}}
& \makecell[c]{20.5$\std{0.4}$ \\ \down{-5.6}}
& \makecell[c]{15.0$\std{0.1}$ \\ \down{-1.8}} \\
\quad \makecell[l]{Group-key only \\ {\scriptsize (Always B+G)}}
& \makecell[c]{74.5$\std{0.5}$ \\ \down{-0.9}}
& \makecell[c]{62.6$\std{0.1}$ \\ \down{-0.2}}
& \makecell[c]{20.9$\std{0.8}$ \\ \down{-5.2}}
& \makecell[c]{14.8$\std{0.1}$ \\ \down{-2.0}} \\
\quad \makecell[l]{Task-key only \\ {\scriptsize (Always B+G+T)}}
& \makecell[c]{75.1$\std{0.2}$ \\ \down{-0.3}}
& \makecell[c]{62.7$\std{0.1}$ \\ \down{-0.1}}
& \makecell[c]{26.1$\std{0.3}$ \\ \same}
& \makecell[c]{16.0$\std{0.2}$ \\ \down{-0.8}} \\
\bottomrule
\bottomrule
\end{tabular}
\caption{Ablation study of HiVe on MRQA and summarization. The evaluation covers the three core components: residual decomposition, hierarchical structure, and V-MoE routing. Performance changes relative to the full model are shown in parentheses.}
\vspace{-0.4cm}
\label{table:ablation}
\end{table}

\subsubsection{Ablation Study}

Table~\ref{table:ablation} presents ablation studies on the core components of \algname{}, including residual decomposition, hierarchical structure, and V-MoE routing. Additional ablation results are provided in Appendix~\ref{appendix:ablation}.

\textbf{Residual Decomposition.}
Removing residual decomposition (\textit{w/o Residual Decomposition}) degrades performance in both in-domain and out-of-domain datasets, with larger drops in OOD settings. This suggests that, without decomposition, shared and task-specific representations become entangled, making it difficult to preserve shared representations across hierarchy levels. Consequently, each level may fail to capture its intended role.

\textbf{Hierarchical Architecture.}
Variants using only a single hierarchy level (\textit{Base-level only}, \textit{Group-level only}, and \textit{Task-level only}) consistently underperform the full model. In particular, \textit{Task-level only} remains relatively strong in in-domain settings but shows larger degradation under domain shifts, whereas \textit{Base-level only} and \textit{Group-level only} exhibit lower overall performance. These results suggest that the proposed Base–Group–Task hierarchy effectively balances shared and task-specific representations across hierarchy levels.

\textbf{Vertical Mixture of Experts.}
Static routing variants using only a single key type (\textit{Base-key only}, \textit{Group-key only}, and \textit{Task-key only}) generally degrade performance across most settings, particularly on summarization tasks. Although \textit{Task-key only} maintains relatively strong performance, its robustness under domain shifts is reduced. In contrast, \textit{Base-key only} slightly improves performance on MRQA out-of-domain tasks. These findings suggest that the optimal level of specialization may vary depending on the input, supporting the need for multi-level dynamic routing.

\begin{table*}[t]
\centering
\scriptsize
\renewcommand{\arraystretch}{1.2}

\begin{tabularx}{\textwidth}{l X X X}
\toprule
\textbf{Field} 
& \textbf{Case 1} 
& \textbf{Case 2} 
& \textbf{Case 3} \\
\midrule

\textbf{Context}
& Comets are small, icy objects that have very elliptical orbits around the Sun. Their orbits carry them from the outer solar system to the inner solar system, close to the Sun. Early in Earth's history, comets may have brought \colorbox{yellow!60}{water} and other substances to Earth during collisions. Comet tails form the outer layers of ice melt...
& \colorbox{yellow!60}{The Stock Market crash in New York} led people to hoard their money ; as consumption fell , the American economy steadily contracted , 1929 - 32 . Given the close economic links between the two countries , the collapse quickly affected Canada . Added to the woes of the prairies were those of Ontario...
& On June 4, 2014, the NFL announced that the practice of branding Super Bowl games with Roman numerals, a practice established at Super Bowl V, would be temporarily suspended, and that the game would be named using \colorbox{yellow!60}{Arabic numerals} as Super Bowl 50 as opposed to Super Bowl L. The use of Roman numerals...\\

\midrule

\textbf{Question}
& In the early solar system, comets that struck Earth may have brought in what?
& How did the great depression start in canada?
& What type of numeral did the latest Super Bowl use to designate the game number? \\

\midrule

\textbf{Answer}
& water
& The Stock Market crash in New York
& Arabic numerals \\

\midrule

\textbf{Base}
& \textbf{\underline{water}}
& stock market crash
& Roman \\

\cmidrule(lr){1-4}

\textbf{Base + Group}
& water and other substances
& \textbf{\underline{The Stock Market crash in New York}}
& Roman numerals \\

\cmidrule(lr){1-4}

\textbf{Base + Group + Task}
& water and other substances
& The Stock Market crash in New York led people to hoard their money ; as...
& \textbf{\underline{Arabic numerals}} \\

\bottomrule
\end{tabularx}

\caption{
Case study of progressively specialized prompt compositions in \algname{}.
Underlined predictions indicate the specialization level selected by \algname{}.
The yellow-highlighted spans denote the ground-truth answers.
}
\label{tab:case_study}
\end{table*}

\subsection{In-Depth Analysis}

\subsubsection{Case Study}

We qualitatively analyze the effectiveness of V-MoE routing by comparing predictions generated with progressively expanded hierarchy levels for the same input. As shown in Table~\ref{tab:case_study}, different hierarchy levels produce different degrees of semantic specificity. In some cases, the Base-level predictor generates concise and generalized answers, while the Task-level predictor captures finer contextual details. The Group-level predictor often exhibits an intermediate level of specialization, avoiding both overly generic and excessively verbose predictions. Notably, \algname{} tends to select different hierarchy levels depending on the input characteristics, producing the most appropriate prediction for each case. For example, in Case 1, only the Base-level predictor correctly outputs the concise answer ``water,'' whereas the Group- and Task-level predictors include unnecessary contextual spans. In Case 2, the Group-level predictor produces the most appropriate level of specificity, while the Base-level predictor is overly generic and the Task-level predictor generates an excessively verbose response. Finally, in Case 3, only the Task-level predictor successfully captures the subtle distinction between Roman numerals and Arabic numerals.

\begin{figure*}[t]
    \centering
    \includegraphics[width=\textwidth]{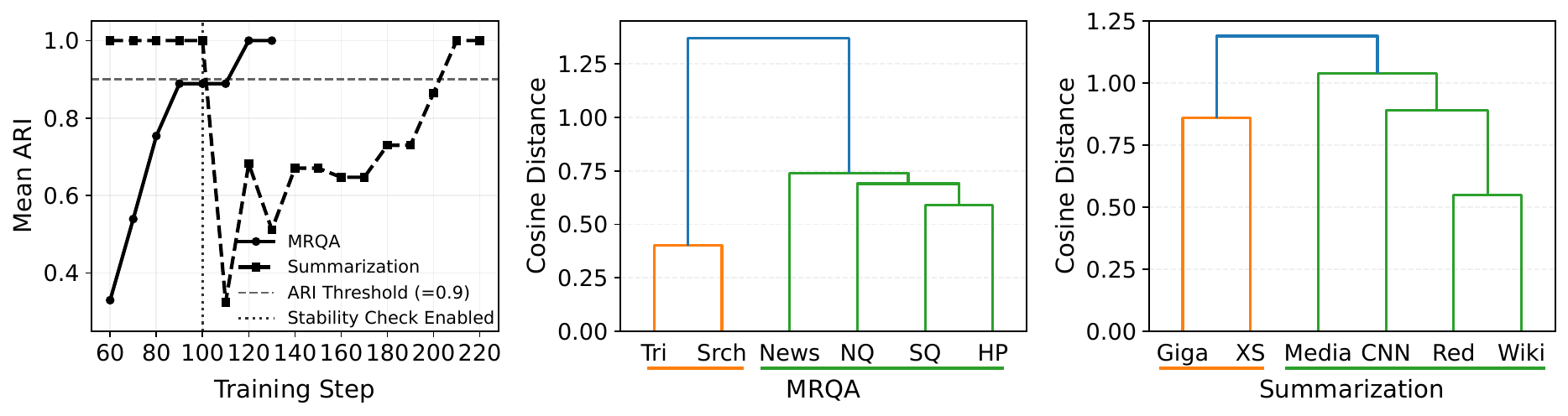}

    \caption{
    Grouping stability during training. The left figure shows the mean ARI across training steps, and the right figure shows HAC-based grouping results, where underlined task names indicate the same group.
    }

    \label{fig:ari}
\end{figure*}

\subsubsection{Grouping Analysis}

Figure~\ref{fig:ari} shows that task grouping gradually stabilizes during training. The mean ARI score consistently increases and eventually satisfies the stability criterion for both MRQA and summarization. To avoid unreliable early-stage group formation, stability checks are activated only after the warm-up stage, and measurement begins only after the rolling window is fully populated. The HAC visualizations further show that semantically related tasks form coherent groups, such as Tri and Srch in MRQA, and Giga and XS in summarization. Although the overall grouping structure remains stable, minor variations are occasionally observed across different hyperparameter settings.

\subsubsection{Routing Analysis}

\begin{figure}[t]
    \centering
    \includegraphics[width=\columnwidth]{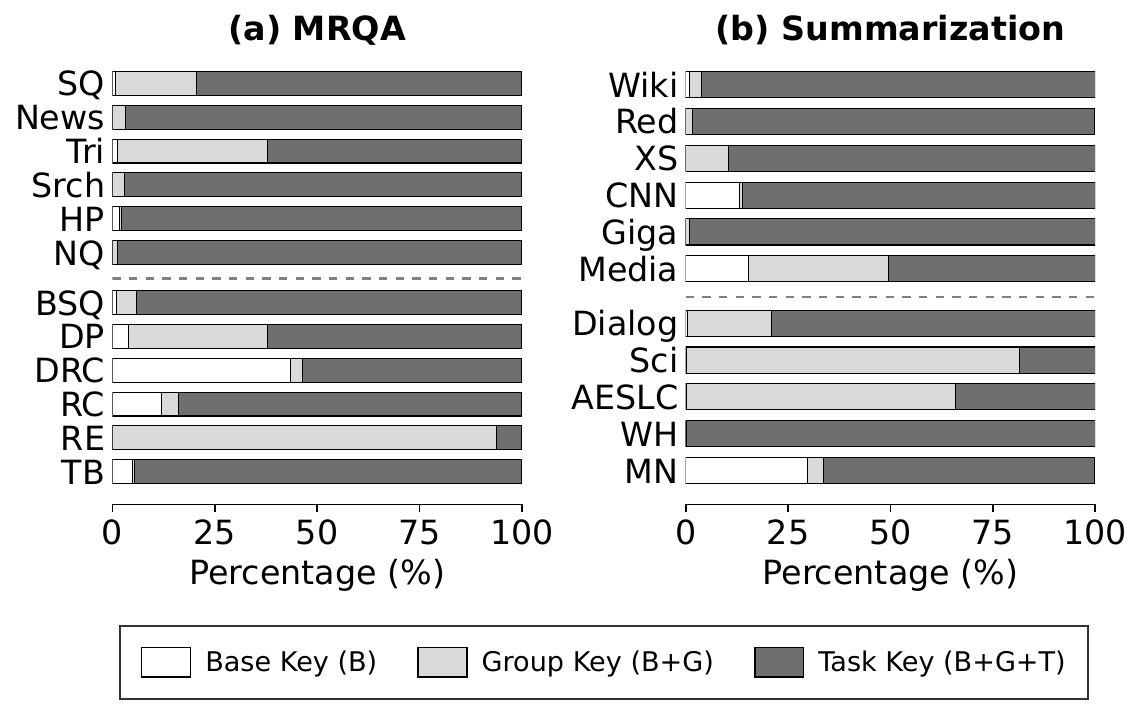}

    \caption{
        Routing key level selection rates across datasets. Different tasks exhibit distinct preferences for hierarchical routing levels.
    }

    \label{fig:routing_distribution}

    \vspace{-1.0em}
\end{figure}

Figure~\ref{fig:routing_distribution} presents the routing key selection distribution across hierarchy levels. Overall, different tasks exhibit distinct routing preferences across hierarchy levels. For example, RE strongly favors Group-level routing, while DRC shows increased reliance on Base-level routing. In contrast, several tasks more frequently utilize Task-level routing, indicating that the preferred level of specialization varies depending on the task and domain characteristics. These results suggest that using a fixed specialization level may be suboptimal for multi-task adaptation, motivating adaptive routing over hierarchical prompts.

\subsubsection{Model Hyperparameter Sensitivity}
\begin{figure}[t]
    \centering
    \includegraphics[width=\columnwidth]{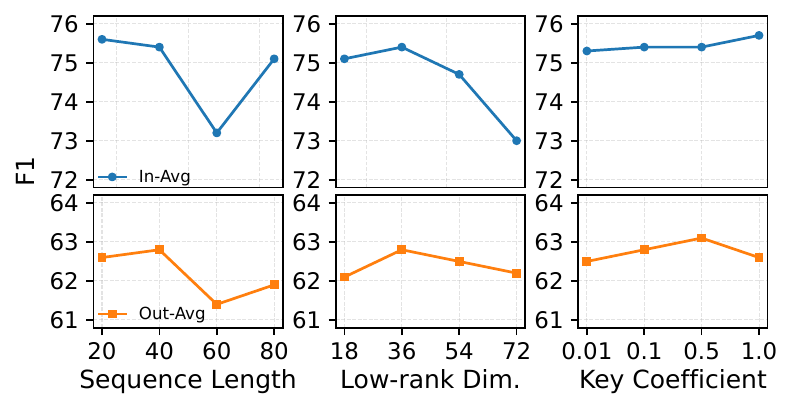}

    \caption{
    Hyperparameter sensitivity analysis of \algname{} on MRQA across different configurations.
    From left to right, the figures illustrate the effects of prompt length, low-rank dimension, and key loss coefficient on in-domain and out-of-domain performance.
    }

    \label{fig:mean_hyper}
\end{figure}

Figure~\ref{fig:mean_hyper} shows the effects of three main hyperparameters used in \algname{}.

\textbf{Prompt Length.} We analyze the effect of prompt length (20, 40, 60, and 80). Overall, the most stable performance is observed at shorter prompt lengths (20--40), while longer lengths tend to degrade performance. In particular, both in-domain and out-of-domain performance noticeably decrease at length 60. Although partial recovery is observed at length 80, the overall performance remains lower than that of shorter prompt lengths. These results suggest that large prompt spaces do not necessarily yield better results.

\textbf{Low-Rank Dimension.} We analyze the effect of low-rank dimensionality (18, 36, 54, and 72). Performance consistently peaks at dimension 36 across both in-domain and out-of-domain settings. Larger dimensions gradually degrade performance, with a more pronounced decline in the in-domain setting. These results suggest that increasing the low-rank dimension beyond a moderate size does not provide additional performance gains.

\textbf{Key Coefficient.} We also evaluate the effect of the key loss coefficient $\lambda$ on MRQA, which controls the strength of routing supervision. Small coefficients lead to insufficient routing optimization, while excessively large coefficients occasionally degrade overall generalization performance, possibly because the shared projection matrix $W$ used by both prompts and routing keys becomes more biased toward routing-specific features. Overall, moderate coefficients (0.1--0.5) provide the most stable performance.

\begin{table}[t]
\centering
\footnotesize
\setlength{\tabcolsep}{6pt}
\renewcommand{\arraystretch}{1.08}

\begin{tabular}{lcccc}
\toprule
\textbf{Setting} & \textbf{Step} & \textbf{Part.} & \textbf{In} & \textbf{OOD} \\
\midrule
Default       & 220 & A & 26.5 & 16.6 \\
\midrule
$K{=}3$       & 200 & A & 26.4 & 16.7 \\
$K{=}7$       & 230 & A & 26.5 & 17.0 \\
$N{=}5$       & 185 & A & 26.5 & 16.2 \\
$N{=}20$      & 280 & A & 26.3 & 16.0 \\
$\eta{=}0.7$  & 190 & A & 26.4 & 16.2 \\
$\eta{=}0.8$  & 210 & A & 26.5 & 16.6 \\
$R{=}1$       & 100 & B & 25.6 & 16.1 \\
$R{=}3$       & 230 & A & 26.5 & 17.0 \\
\bottomrule
\end{tabular}

\caption{Grouping hyperparameter sensitivity on summarization using a single seed. Partitions A and B denote \{Giga, XS\}/\{Red, Wiki, CNN, Media\} and \{Red, Wiki, CNN\}/\{Giga, XS, Media\}, respectively.}
\label{tab:grouping_hyper}
\end{table}

\begin{table}[t]
\centering
\footnotesize
\setlength{\tabcolsep}{4pt}
\renewcommand{\arraystretch}{1.0}

\begin{tabular}{llcccc}
\toprule
\textbf{Benchmark} & \textbf{Seed} & \textbf{Step} & \textbf{Part.} & \textbf{In} & \textbf{OOD} \\
\midrule
MRQA  & 1 & 130 & A & 75.3 & 62.7 \\
MRQA  & 2 & 110 & A & 75.7 & 62.6 \\
MRQA  & 3 & 160 & A & 75.2 & 63.2 \\
\midrule
Summ. & 1 & 220 & B & 26.5 & 16.6 \\
Summ. & 2 & 280 & B & 26.4 & 16.7 \\
Summ. & 3 & 110 & C & 25.4 & 17.1 \\
\bottomrule
\end{tabular}

\caption{Seed sensitivity of task grouping. Partitions A, B, and C denote \{Tri, Srch\}/\{SQ, News, HP, NQ\}, \{Giga, XS\}/\{Red, Wiki, CNN, Media\}, and \{CNN, Giga, Media\}/\{Wiki, Red, XS\}, respectively.}
\label{tab:grouping_seed}
\end{table}

\subsubsection{Grouping Sensitivity}

We further analyze the sensitivity of task grouping to its hyperparameters and random initialization.

\textbf{Hyperparameter Sensitivity.}
Table~\ref{tab:grouping_hyper} reports the effects of the grouping hyperparameters on summarization by varying the window size $K$, check interval $N$, ARI threshold $\eta$, and patience $R$. Across the tested ranges, these hyperparameters mainly affect when grouping is finalized, while the resulting partition and downstream performance remain largely stable. The main exception is $R{=}1$, which finalizes grouping earlier, produces a different partition, and lowers in-domain performance. This suggests that grouping should be finalized only after stability is confirmed across multiple checks.

\textbf{Random-Seed Sensitivity.}
Table~\ref{tab:grouping_seed} evaluates the sensitivity of task grouping to random seeds. On MRQA, all seeds recover the same partition and exhibit similar in-domain and OOD performance. On summarization, two seeds yield the same partition, whereas one seed produces a different grouping with lower in-domain but slightly higher OOD performance. Thus, seed-dependent partition variation does not consistently improve or degrade performance in either the in-domain or OOD setting.

\section{Conclusion}

We proposed \algname{}, a hierarchical prompt tuning framework that adaptively controls the level of prompt specialization for each input through vertical mixture-of-experts (V-MoE) routing. By combining stability-driven task grouping and residual prompt decomposition, \algname{} effectively disentangles shared and task-specific knowledge across hierarchical levels. Beyond parameter sharing, the learned hierarchy also serves as an input-dependent routing space for controlling the degree of specialization required by each input. Experimental results on question answering and summarization benchmarks demonstrate that \algname{} consistently outperforms strong prompt tuning baselines in both in-domain and out-of-domain settings while maintaining parameter efficiency. Overall, our findings suggest that dynamically controlling prompt specialization provides an effective direction for multitask learning. Future work could explore soft or top-$k$ routing as alternatives to the current hard-routing scheme for more flexible specialization across hierarchical levels.
\section*{Limitations}

\algname{} relies on stability-driven task grouping to construct hierarchical prompts. Since the group structure emerges from learned task prompt representations, it can be influenced by factors such as random seeds, prompt length, low-rank dimensionality, and grouping hyperparameters. As a result, some configurations may finalize grouping prematurely, leading to different group structures and reduced downstream performance. Furthermore, HiVe is evaluated on within-task distribution shifts, where the task objective remains unchanged while the input distribution varies, while generalization across task types remains outside the scope of this study. Adaptive routing can also mitigate over- or under-specialization for unseen inputs, but cannot capture patterns beyond the learned hierarchy, leaving gaps under substantial distribution shifts.

\smallskip
\section*{Acknowledgments}
This work was supported by the Institute of Information \& Communications Technology Planning \& Evaluation (IITP) grants funded by the Korea government(MSIT) (No. RS-2026-25507543 (40\%), No. IITP-2026-RS-2025-02304828 (40\%), No. IITP-2026-RS-2020-II201819 (10\%), and No. RS-2026-25585074 (10\%)).

\bibliography{custom}

@article{wang-etal-2025-parameter,
    author = "Wang, Luping and Chen, Sheng and Jiang, Linnan and Pan, Shu and Cai, Runze and Yang, Sen and Yang, Fei",
    title = "Parameter-efficient fine-tuning in large language models: A survey of methodologies",
    journal = "Artificial Intelligence Review",
    volume = "58",
    number = "8",
    year = "2025",
    url = "https://doi.org/10.1007/s10462-025-11236-4",
    doi = "10.1007/s10462-025-11236-4",
    pages = "227"
}

@inproceedings{lester-etal-2021-power,
    author = "Lester, Brian and Al-Rfou, Rami and Constant, Noah",
    title = "The power of scale for parameter-efficient prompt tuning",
    booktitle = "Proceedings of the 2021 Conference on Empirical Methods in Natural Language Processing",
    year = "2021",
    address = "Online and Punta Cana, Dominican Republic",
    publisher = "Association for Computational Linguistics",
    url = "https://aclanthology.org/2021.emnlp-main.243",
    doi = "10.18653/v1/2021.emnlp-main.243",
    pages = "3045--3059"
}

@inproceedings{dun-etal-2025-sweeping,
    author = "Dun, Chen and Garcia, Mirian Hipolito and Zheng, Guoqing and Awadallah, Ahmed H. and Sim, Robert and Kyrillidis, Anastasios",
    title = "Sweeping heterogeneity with smart mops: Mixture of prompts for {LLM} task adaptation",
    booktitle = "Proceedings of the {AAAI} Conference on Artificial Intelligence",
    volume = "39",
    number = "16",
    year = "2025",
    address = "Philadelphia, USA",
    publisher = "AAAI Press",
    url = "https://arxiv.org/abs/2310.02842",
    pages = "16426--16434"
}

@inproceedings{choi-etal-2023-smop,
    author = "Choi, Joon-Young and Kim, Junho and Park, Jun-Hyung and Mok, Wing-Lam and Lee, SangKeun",
    title = "{SMoP}: Towards efficient and effective prompt tuning with sparse mixture-of-prompts",
    booktitle = "Proceedings of the 2023 Conference on Empirical Methods in Natural Language Processing",
    year = "2023",
    address = "Singapore",
    publisher = "Association for Computational Linguistics",
    url = "https://aclanthology.org/2023.emnlp-main.884",
    doi = "10.18653/v1/2023.emnlp-main.884",
    pages = "14306--14316"
}

@inproceedings{li-etal-2025-ptmoe,
    author = "Li, Zongqian and Su, Yixuan and Collier, Nigel",
    title = "{PT-MoE}: An efficient finetuning framework for integrating mixture-of-experts into prompt tuning",
    booktitle = "Advances in Neural Information Processing Systems",
    volume = "38",
    year = "2025",
    address = "San Diego, California",
    publisher = "Curran Associates, Inc.",
    url = "https://proceedings.neurips.cc/paper_files/paper/2025/hash/77f33e8bd80345de4aea8554bbe5a4da-Abstract-Conference.html",
    pages = "83240--83261"
}

@inproceedings{wang-etal-2022-hpt,
    author = "Wang, Zihan and Wang, Peiyi and Liu, Tianyu and Lin, Binghuai and Cao, Yunbo and Sui, Zhifang and Wang, Houfeng",
    title = "{HPT}: Hierarchy-aware prompt tuning for hierarchical text classification",
    booktitle = "Proceedings of the 2022 Conference on Empirical Methods in Natural Language Processing",
    year = "2022",
    address = "Abu Dhabi, United Arab Emirates",
    publisher = "Association for Computational Linguistics",
    url = "https://aclanthology.org/2022.emnlp-main.246",
    doi = "10.18653/v1/2022.emnlp-main.246",
    pages = "3740--3751"
}

@inproceedings{chen-etal-2023-mprompt,
    author = "Chen, Guoxin and Qian, Yiming and Wang, Bowen and Li, Liangzhi",
    title = "{MPrompt}: Exploring multi-level prompt tuning for machine reading comprehension",
    booktitle = "Findings of the Association for Computational Linguistics: EMNLP 2023",
    year = "2023",
    address = "Singapore",
    publisher = "Association for Computational Linguistics",
    url = "https://aclanthology.org/2023.findings-emnlp.343",
    doi = "10.18653/v1/2023.findings-emnlp.343",
    pages = "5163--5175"
}

@inproceedings{liu-etal-2023-hierarchical,
    author = "Liu, Yajing and Lu, Yuning and Liu, Hao and An, Yaozu and Xu, Zhuoran and Yao, Zhuokun and Zhang, Baofeng and Xiong, Zhiwei and Gui, Chenguang",
    title = "Hierarchical prompt learning for multi-task learning",
    booktitle = "Proceedings of the {IEEE/CVF} Conference on Computer Vision and Pattern Recognition",
    year = "2023",
    address = "Vancouver, Canada",
    publisher = "IEEE",
    url = "https://openaccess.thecvf.com/content/CVPR2023/html/Liu_Hierarchical_Prompt_Learning_for_Multi-Task_Learning_CVPR_2023_paper.html",
    pages = "10888--10898"
}

@inproceedings{shazeer-etal-2017-outrageously,
    author = "Shazeer, Noam and Mirhoseini, Azalia and Maziarz, Krzysztof and Davis, Andy and Le, Quoc and Hinton, Geoffrey and Dean, Jeff",
    title = "Outrageously large neural networks: The sparsely-gated mixture-of-experts layer",
    booktitle = "Proceedings of the 5th International Conference on Learning Representations ({ICLR})",
    year = "2017",
    address = "Toulon, France",
    publisher = "OpenReview.net",
    url = "https://openreview.net/forum?id=B1ckMDqlg"
}

@inproceedings{wang-etal-2022-adamix,
    author = "Wang, Yaqing and Agarwal, Sahaj and Mukherjee, Subhabrata and Liu, Xiaodong and Gao, Jing and Awadallah, Ahmed Hassan and Gao, Jianfeng",
    title = "{A}da{M}ix: Mixture-of-adaptations for parameter-efficient model tuning",
    booktitle = "Proceedings of the 2022 Conference on Empirical Methods in Natural Language Processing",
    year = "2022",
    address = "Abu Dhabi, United Arab Emirates",
    publisher = "Association for Computational Linguistics",
    url = "https://aclanthology.org/2022.emnlp-main.388",
    doi = "10.18653/v1/2022.emnlp-main.388",
    pages = "5744--5760"
}

@inproceedings{tian-etal-2024-hydralora,
    author = "Tian, Chunlin and Shi, Zhan and Guo, Zhijiang and Li, Li and Xu, Chengzhong",
    title = "{H}ydra{L}o{R}A: An asymmetric {L}o{R}A architecture for efficient fine-tuning",
    booktitle = "Advances in Neural Information Processing Systems",
    volume = "37",
    year = "2024",
    address = "Vancouver, Canada",
    publisher = "Curran Associates, Inc.",
    url = "https://proceedings.neurips.cc/paper_files/paper/2024/hash/123fd8a56501194823c8e0dca00733df-Abstract-Conference.html"
}

@inproceedings{hu-etal-2022-lora,
    author = "Hu, Edward J. and Shen, Yelong and Wallis, Phillip and Allen-Zhu, Zeyuan and Li, Yuanzhi and Wang, Shean Gold and Wang, Lu and Chen, Weizhu",
    title = "{L}o{R}A: Low-rank adaptation of large language models",
    booktitle = "Proceedings of the 10th International Conference on Learning Representations ({ICLR})",
    year = "2022",
    address = "Online",
    publisher = "OpenReview.net",
    url = "https://openreview.net/forum?id=nZeVKeeFYf9"
}

@inproceedings{asai-etal-2022-attempt,
    author = "Asai, Akari and Salehi, Mohammadreza and Peters, Matthew and Hajishirzi, Hannaneh",
    title = "{ATTEMPT}: Parameter-efficient multi-task tuning via attentional mixtures of soft prompts",
    booktitle = "Proceedings of the 2022 Conference on Empirical Methods in Natural Language Processing",
    year = "2022",
    address = "Abu Dhabi, United Arab Emirates",
    publisher = "Association for Computational Linguistics",
    url = "https://aclanthology.org/2022.emnlp-main.446/",
    doi = "10.18653/v1/2022.emnlp-main.446",
    pages = "6655--6672"
}

@inproceedings{bingel-etal-2017-identifying,
    author = "Bingel, Joachim and S{\o}gaard, Anders",
    title = "Identifying beneficial task relations for multi-task learning in deep neural networks",
    booktitle = "Proceedings of the 15th Conference of the European Chapter of the Association for Computational Linguistics: Volume 2, Short Papers",
    year = "2017",
    address = "Valencia, Spain",
    publisher = "Association for Computational Linguistics",
    url = "https://aclanthology.org/E17-2026/",
    pages = "164--169"
}

@inproceedings{ni-etal-2023-aggregating,
    author = "Ni, Jingwei and Jin, Zhijing and Wang, Qian and Sachan, Mrinmaya and Leippold, Markus",
    title = "When Does Aggregating Multiple Skills with Multi-Task Learning Work? A Case Study in Financial {NLP}",
    booktitle = "Proceedings of the 61st Annual Meeting of the Association for Computational Linguistics (Volume 1: Long Papers)",
    year = "2023",
    address = "Toronto, Canada",
    publisher = "Association for Computational Linguistics",
    url = "https://aclanthology.org/2023.acl-long.412/",
    doi = "10.18653/v1/2023.acl-long.412",
    pages = "7465--7488"
}

@inproceedings{kumar-etal-2012-learning,
    author = "Kumar, Abhishek and Daum{\'e} III, Hal",
    title = "Learning task grouping and overlap in multi-task learning",
    booktitle = "Proceedings of the 29th International Conference on Machine Learning",
    year = "2012",
    address = "Edinburgh, Scotland, UK",
    publisher = "PMLR",
    url = "https://arxiv.org/abs/1206.6417",
    pages = "1383--1390"
}

@inproceedings{fifty-etal-2021-efficiently,
    author = "Fifty, Chris and Amid, Ehsan and Zhao, Zhe and Yu, Tianhe and Anil, Rohan and Finn, Chelsea",
    title = "Efficiently identifying task groupings for multi-task learning",
    booktitle = "Advances in Neural Information Processing Systems",
    volume = "34",
    year = "2021",
    address = "Online",
    publisher = "Curran Associates, Inc.",
    url = "https://proceedings.neurips.cc/paper/2021/hash/e77910ebb93b511588557806310f78f1-Abstract.html",
    pages = "27503--27516"
}

@inproceedings{zhang-etal-2025-dynamic,
    author = "Zhang, Peiyi and Zhang, Richong and Nie, Zhijie and Wang, Ziqiao",
    title = "{D}ynamic task vector grouping for efficient multi-task prompt tuning",
    booktitle = "Findings of the Association for Computational Linguistics: ACL 2025",
    year = "2025",
    address = "Vienna, Austria",
    publisher = "Association for Computational Linguistics",
    url = "https://aclanthology.org/2025.findings-acl.1374/",
    doi = "10.18653/v1/2025.findings-acl.1374",
    pages = "26805--26821"
}

@article{hubert-etal-1985-comparing,
    author = "Hubert, Lawrence and Arabie, Phipps",
    title = "Comparing partitions",
    journal = "Journal of Classification",
    volume = "2",
    number = "1",
    year = "1985",
    address = "New York, USA",
    publisher = "Springer-Verlag",
    url = "https://link.springer.com/article/10.1007/BF01908075",
    doi = "10.1007/BF01908075",
    pages = "193--218"
}

@inproceedings{fisch-etal-2019-mrqa,
    author = "Fisch, Adam and Talmor, Alon and Jia, Robin and Seo, Minjoon and Choi, Eunsol and Chen, Danqi",
    title = "{MRQA} 2019 shared task: Evaluating generalization in reading comprehension",
    booktitle = "Proceedings of the 2nd Workshop on Machine Reading for Question Answering",
    year = "2019",
    address = "Hong Kong, China",
    publisher = "Association for Computational Linguistics",
    url = "https://aclanthology.org/D19-5801/",
    doi = "10.18653/v1/D19-5801",
    pages = "1--13"
}

@inproceedings{xiao-etal-2023-decomposed,
    author = "Xiao, Yao and Xu, Lu and Li, Jiaxi and Lu, Wei and Li, Xiaoli",
    title = "Decomposed Prompt Tuning via Low-Rank Reparameterization",
    booktitle = "Findings of the Association for Computational Linguistics: EMNLP 2023",
    year = "2023",
    address = "Singapore",
    publisher = "Association for Computational Linguistics",
    url = "https://aclanthology.org/2023.findings-emnlp.890/",
    doi = "10.18653/v1/2023.findings-emnlp.890",
    pages = "13335--13347"
}

@inproceedings{rajpurkar-etal-2016-squad,
    title = "{SQ}u{AD}: 100,000+ Questions for Machine Comprehension of Text",
    author = "Rajpurkar, Pranav  and
      Zhang, Jian  and
      Lopyrev, Konstantin  and
      Liang, Percy",
    editor = "Su, Jian  and
      Duh, Kevin  and
      Carreras, Xavier",
    booktitle = "Proceedings of the 2016 Conference on Empirical Methods in Natural Language Processing",
    month = nov,
    year = "2016",
    address = "Austin, Texas",
    publisher = "Association for Computational Linguistics",
    url = "https://aclanthology.org/D16-1264/",
    doi = "10.18653/v1/D16-1264",
    pages = "2383--2392"
}

@inproceedings{trischler-etal-2017-newsqa,
    title = "{N}ews{QA}: A Machine Comprehension Dataset",
    author = "Trischler, Adam  and
      Wang, Tong  and
      Yuan, Xingdi  and
      Harris, Justin  and
      Sordoni, Alessandro  and
      Bachman, Philip  and
      Suleman, Kaheer",
    editor = "Blunsom, Phil  and
      Bordes, Antoine  and
      Cho, Kyunghyun  and
      Cohen, Shay  and
      Dyer, Chris  and
      Grefenstette, Edward  and
      Hermann, Karl Moritz  and
      Rimell, Laura  and
      Weston, Jason  and
      Yih, Scott",
    booktitle = "Proceedings of the 2nd Workshop on Representation Learning for {NLP}",
    month = aug,
    year = "2017",
    address = "Vancouver, Canada",
    publisher = "Association for Computational Linguistics",
    url = "https://aclanthology.org/W17-2623/",
    doi = "10.18653/v1/W17-2623",
    pages = "191--200"
}

@inproceedings{joshi-etal-2017-triviaqa,
    title = "{T}rivia{QA}: A Large Scale Distantly Supervised Challenge Dataset for Reading Comprehension",
    author = "Joshi, Mandar  and
      Choi, Eunsol  and
      Weld, Daniel  and
      Zettlemoyer, Luke",
    editor = "Barzilay, Regina  and
      Kan, Min-Yen",
    booktitle = "Proceedings of the 55th Annual Meeting of the Association for Computational Linguistics (Volume 1: Long Papers)",
    month = jul,
    year = "2017",
    address = "Vancouver, Canada",
    publisher = "Association for Computational Linguistics",
    url = "https://aclanthology.org/P17-1147/",
    doi = "10.18653/v1/P17-1147",
    pages = "1601--1611"
}

@article{dunn-etal-2017-searchqa,
    title = "{S}earch{QA}: A New Q\&A Dataset Augmented with Context from a Search Engine",
    author = "Dunn, Matthew  and
      Sagun, Levent  and
      Higgins, Mike  and
      Guney, V. Ugur  and
      Cirik, Volkan  and
      Cho, Kyunghyun",
    journal = "arXiv preprint arXiv:1704.05179",
    year = "2017",
    url = "https://arxiv.org/abs/1704.05179"
}

@inproceedings{yang-etal-2018-hotpotqa,
    title = "{H}otpot{QA}: A Dataset for Diverse, Explainable Multi-hop Question Answering",
    author = "Yang, Zhilin  and
      Qi, Peng  and
      Zhang, Saizheng  and
      Bengio, Yoshua  and
      Cohen, William  and
      Salakhutdinov, Ruslan  and
      Manning, Christopher D.",
    editor = "Riloff, Ellen  and
      Chiang, David  and
      Hockenmaier, Julia  and
      Tsujii, Jun{'}ichi",
    booktitle = "Proceedings of the 2018 Conference on Empirical Methods in Natural Language Processing",
    month = oct # "-" # nov,
    year = "2018",
    address = "Brussels, Belgium",
    publisher = "Association for Computational Linguistics",
    url = "https://aclanthology.org/D18-1259/",
    doi = "10.18653/v1/D18-1259",
    pages = "2369--2380"
}

@article{kwiatkowski-etal-2019-natural,
    title = "Natural Questions: A Benchmark for Question Answering Research",
    author = "Kwiatkowski, Tom  and
      Palomaki, Jennimaria  and
      Redfield, Olivia  and
      Collins, Michael  and
      Parikh, Ankur  and
      Alberti, Chris  and
      Epstein, Danielle  and
      Polosukhin, Illia  and
      Devlin, Jacob  and
      Lee, Kenton  and
      Toutanova, Kristina  and
      Jones, Llion  and
      Kelcey, Matthew  and
      Chang, Ming-Wei  and
      Dai, Andrew M.  and
      Uszkoreit, Jakob  and
      Le, Quoc  and
      Petrov, Slav",
    editor = "Lee, Lillian  and
      Johnson, Mark  and
      Roark, Brian  and
      Nenkova, Ani",
    journal = "Transactions of the Association for Computational Linguistics",
    volume = "7",
    year = "2019",
    address = "Cambridge, MA",
    publisher = "MIT Press",
    url = "https://aclanthology.org/Q19-1026/",
    doi = "10.1162/tacl_a_00276",
    pages = "452--466"
}

@inproceedings{tsatsaronis-etal-2012-bioasq,
    title = "{B}io{ASQ}: A Challenge on Large-Scale Biomedical Semantic Indexing and Question Answering",
    author = "Tsatsaronis, George  and
      Schroeder, Michael  and
      Paliouras, Georgios  and
      Almirantis, Yannis  and
      Androutsopoulos, Ion  and
      Gaussier, Eric  and
      Gallinari, Patrick  and
      Artieres, Thierry  and
      Alvers, Michael R.  and
      Zschunke, Matthias  and
      Ngonga Ngomo, Axel-Cyrille",
    booktitle = "Proceedings of the AAAI Fall Symposium on Information Retrieval and Knowledge Discovery in Biomedical Text",
    year = "2012",
    address = "Arlington, Virginia",
    publisher = "AAAI Press",
    pages = "92--98",
    url = "https://aaai.org/papers/05600-5600-bioasq-a-challenge-on-large-scale-biomedical-semantic-indexing-and-question-answering/"
}

@inproceedings{dua-etal-2019-drop,
    title = "{DROP}: A Reading Comprehension Benchmark Requiring Discrete Reasoning Over Paragraphs",
    author = "Dua, Dheeru  and
      Wang, Yizhong  and
      Dasigi, Pradeep  and
      Stanovsky, Gabriel  and
      Singh, Sameer  and
      Gardner, Matt",
    editor = "Burstein, Jill  and
      Doran, Christy  and
      Solorio, Thamar",
    booktitle = "Proceedings of the 2019 Conference of the North {A}merican Chapter of the Association for Computational Linguistics: Human Language Technologies, Volume 1 (Long and Short Papers)",
    month = jun,
    year = "2019",
    address = "Minneapolis, Minnesota",
    publisher = "Association for Computational Linguistics",
    url = "https://aclanthology.org/N19-1246/",
    doi = "10.18653/v1/N19-1246",
    pages = "2368--2378"
}

@inproceedings{saha-etal-2018-duorc,
    title = "{D}uo{RC}: Towards Complex Language Understanding with Paraphrased Reading Comprehension",
    author = "Saha, Amrita  and
      Aralikatte, Rahul  and
      Khapra, Mitesh M.  and
      Sankaranarayanan, Karthik",
    editor = "Gurevych, Iryna  and
      Miyao, Yusuke",
    booktitle = "Proceedings of the 56th Annual Meeting of the Association for Computational Linguistics (Volume 1: Long Papers)",
    month = jul,
    year = "2018",
    address = "Melbourne, Australia",
    publisher = "Association for Computational Linguistics",
    url = "https://aclanthology.org/P18-1156/",
    doi = "10.18653/v1/P18-1156",
    pages = "1683--1693"
}

@inproceedings{lai-etal-2017-race,
    title = "{RACE}: Large-scale {R}e{A}ding Comprehension Dataset From Examinations",
    author = "Lai, Guokun  and
      Xie, Qizhe  and
      Liu, Hanxiao  and
      Yang, Yiming  and
      Hovy, Eduard",
    editor = "Palmer, Martha  and
      Hwa, Rebecca  and
      Riedel, Sebastian",
    booktitle = "Proceedings of the 2017 Conference on Empirical Methods in Natural Language Processing",
    month = sep,
    year = "2017",
    address = "Copenhagen, Denmark",
    publisher = "Association for Computational Linguistics",
    url = "https://aclanthology.org/D17-1082/",
    doi = "10.18653/v1/D17-1082",
    pages = "785--794"
}

@inproceedings{levy-etal-2017-zero,
    title = "Zero-Shot Relation Extraction via Reading Comprehension",
    author = "Levy, Omer  and
      Seo, Minjoon  and
      Choi, Eunsol  and
      Zettlemoyer, Luke",
    editor = "Levy, Roger  and
      Specia, Lucia",
    booktitle = "Proceedings of the 21st Conference on Computational Natural Language Learning ({C}o{NLL} 2017)",
    month = aug,
    year = "2017",
    address = "Vancouver, Canada",
    publisher = "Association for Computational Linguistics",
    url = "https://aclanthology.org/K17-1034/",
    doi = "10.18653/v1/K17-1034",
    pages = "333--342"
}

@inproceedings{kembhavi-etal-2017-are,
    author = "Kembhavi, Aniruddha  and
      Seo, Minjoon  and
      Schwenk, Dustin  and
      Choi, Jonghyun  and
      Farhadi, Ali  and
      Hajishirzi, Hannaneh",
    title = "Are You Smarter Than a Sixth Grader? Textbook Question Answering for Multimodal Machine Comprehension",
    booktitle = "Proceedings of the IEEE Conference on Computer Vision and Pattern Recognition ({CVPR})",
    month = jul,
    year = "2017",
    address = "Honolulu, Hawaii",
    publisher = "IEEE",
    pages = "4999--5007",
    url = "https://openaccess.thecvf.com/content_cvpr_2017/html/Kembhavi_Are_You_Smarter_CVPR_2017_paper.html"
}

@inproceedings{ladhak-etal-2020-wikilingua,
    title = "{W}iki{L}ingua: A New Benchmark Dataset for Cross-Lingual Abstractive Summarization",
    author = "Ladhak, Faisal  and
      Durmus, Esin  and
      Cardie, Claire  and
      McKeown, Kathleen",
    editor = "Cohn, Trevor  and
      He, Yulan  and
      Liu, Yang",
    booktitle = "Findings of the Association for Computational Linguistics: EMNLP 2020",
    month = nov,
    year = "2020",
    address = "Online",
    publisher = "Association for Computational Linguistics",
    url = "https://aclanthology.org/2020.findings-emnlp.360/",
    doi = "10.18653/v1/2020.findings-emnlp.360",
    pages = "4034--4048"
}

@inproceedings{volske-etal-2017-tl,
    title = "{TL};{DR}: Mining {R}eddit to Learn Automatic Summarization",
    author = {V{\"o}lske, Michael  and
      Potthast, Martin  and
      Syed, Shahbaz  and
      Stein, Benno},
    editor = "Wang, Lu  and
      Cheung, Jackie Chi Kit  and
      Carenini, Giuseppe  and
      Liu, Fei",
    booktitle = "Proceedings of the Workshop on New Frontiers in Summarization",
    month = sep,
    year = "2017",
    address = "Copenhagen, Denmark",
    publisher = "Association for Computational Linguistics",
    url = "https://aclanthology.org/W17-4508/",
    doi = "10.18653/v1/W17-4508",
    pages = "59--63"
}

@inproceedings{narayan-etal-2018-dont,
    title = "Don{'}t Give Me the Details, Just the Summary! Topic-Aware Convolutional Neural Networks for Extreme Summarization",
    author = "Narayan, Shashi  and
      Cohen, Shay B.  and
      Lapata, Mirella",
    editor = "Riloff, Ellen  and
      Chiang, David  and
      Hockenmaier, Julia  and
      Tsujii, Jun{'}ichi",
    booktitle = "Proceedings of the 2018 Conference on Empirical Methods in Natural Language Processing",
    month = oct # "-" # nov,
    year = "2018",
    address = "Brussels, Belgium",
    publisher = "Association for Computational Linguistics",
    url = "https://aclanthology.org/D18-1206/",
    doi = "10.18653/v1/D18-1206",
    pages = "1797--1807"
}

@inproceedings{hermann-etal-2015-teaching,
    author = "Hermann, Karl Moritz and
      Kocisky, Tomas and
      Grefenstette, Edward and
      Espeholt, Lasse and
      Kay, Will and
      Suleyman, Mustafa and
      Blunsom, Phil",
    title = "Teaching Machines to Read and Comprehend",
    booktitle = "Advances in Neural Information Processing Systems",
    editor = "Cortes, Corinna and
      Lawrence, Neil and
      Lee, Daniel and
      Sugiyama, Masashi and
      Garnett, Roman",
    volume = "28",
    year = "2015",
    publisher = "Curran Associates, Inc.",
    url = "https://proceedings.neurips.cc/paper_files/paper/2015/file/afdec7005cc9f14302cd0474fd0f3c96-Paper.pdf"
}

@misc{graff-etal-2003-english,
    author = "Graff, David and Cieri, Christopher",
    title = "English Gigaword",
    year = "2003",
    publisher = "Linguistic Data Consortium",
    address = "Philadelphia, Pennsylvania",
    url = "https://catalog.ldc.upenn.edu/LDC2003T05",
    doi = "10.35111/0z6y-q265",
    note = "LDC Catalog No. LDC2003T05"
}

@inproceedings{zhu-etal-2021-mediasum,
    title = "{M}edia{S}um: A Large-scale Media Interview Dataset for Dialogue Summarization",
    author = "Zhu, Chenguang  and
      Liu, Yang  and
      Mei, Jie  and
      Zeng, Michael",
    editor = "Toutanova, Kristina  and
      Rumshisky, Anna  and
      Zettlemoyer, Luke  and
      Hakkani-Tur, Dilek  and
      Beltagy, Iz  and
      Bethard, Steven  and
      Cotterell, Ryan  and
      Chakraborty, Tanmoy  and
      Zhou, Yichao",
    booktitle = "Proceedings of the 2021 Conference of the North American Chapter of the Association for Computational Linguistics: Human Language Technologies",
    month = jun,
    year = "2021",
    address = "Online",
    publisher = "Association for Computational Linguistics",
    url = "https://aclanthology.org/2021.naacl-main.474/",
    doi = "10.18653/v1/2021.naacl-main.474",
    pages = "5927--5934"
}

@inproceedings{chen-etal-2021-dialogsum,
    title = "{D}ialog{S}um: {A} Real-Life Scenario Dialogue Summarization Dataset",
    author = "Chen, Yulong  and
      Liu, Yang  and
      Chen, Liang  and
      Zhang, Yue",
    editor = "Zong, Chengqing  and
      Xia, Fei  and
      Li, Wenjie  and
      Navigli, Roberto",
    booktitle = "Findings of the Association for Computational Linguistics: ACL-IJCNLP 2021",
    month = aug,
    year = "2021",
    address = "Online",
    publisher = "Association for Computational Linguistics",
    url = "https://aclanthology.org/2021.findings-acl.449/",
    doi = "10.18653/v1/2021.findings-acl.449",
    pages = "5062--5074"
}

@inproceedings{cachola-etal-2020-tldr,
    title = "{TLDR}: Extreme Summarization of Scientific Documents",
    author = "Cachola, Isabel  and
      Lo, Kyle  and
      Cohan, Arman  and
      Weld, Daniel",
    editor = "Cohn, Trevor  and
      He, Yulan  and
      Liu, Yang",
    booktitle = "Findings of the Association for Computational Linguistics: EMNLP 2020",
    month = nov,
    year = "2020",
    address = "Online",
    publisher = "Association for Computational Linguistics",
    url = "https://aclanthology.org/2020.findings-emnlp.428/",
    doi = "10.18653/v1/2020.findings-emnlp.428",
    pages = "4766--4777"
}

@inproceedings{zhang-tetreault-2019-email,
    title = "This Email Could Save Your Life: Introducing the Task of Email Subject Line Generation",
    author = "Zhang, Rui  and
      Tetreault, Joel",
    editor = "Korhonen, Anna  and
      Traum, David  and
      M{\`a}rquez, Llu{\'i}s",
    booktitle = "Proceedings of the 57th Annual Meeting of the Association for Computational Linguistics",
    month = jul,
    year = "2019",
    address = "Florence, Italy",
    publisher = "Association for Computational Linguistics",
    url = "https://aclanthology.org/P19-1043/",
    doi = "10.18653/v1/P19-1043",
    pages = "446--456"
}

@article{koupaee-etal-2018-wikihow,
    title = "{W}iki{H}ow: A Large Scale Text Summarization Dataset",
    author = "Koupaee, Mahnaz and Wang, William Yang",
    journal = "arXiv preprint arXiv:1810.09305",
    year = "2018",
    url = "https://arxiv.org/abs/1810.09305",
    doi = "10.48550/arXiv.1810.09305"
}

@inproceedings{fabbri-etal-2019-multi,
    title = "Multi-News: A Large-Scale Multi-Document Summarization Dataset and Abstractive Hierarchical Model",
    author = "Fabbri, Alexander  and
      Li, Irene  and
      She, Tianwei  and
      Li, Suyi  and
      Radev, Dragomir",
    editor = "Korhonen, Anna  and
      Traum, David  and
      M{\`a}rquez, Llu{\'i}s",
    booktitle = "Proceedings of the 57th Annual Meeting of the Association for Computational Linguistics",
    month = jul,
    year = "2019",
    address = "Florence, Italy",
    publisher = "Association for Computational Linguistics",
    url = "https://aclanthology.org/P19-1102/",
    doi = "10.18653/v1/P19-1102",
    pages = "1074--1084"
}

@inproceedings{kim-etal-2024-adapromptcl,
    title = "One Size Fits All for Semantic Shifts: Adaptive Prompt Tuning for Continual Learning",
    author = "Kim, Doyoung  and
    Yoon, Susik  and
    Park, Dongmin  and
    Lee, Youngjun  and
    Song, Hwanjun  and
    Bang, Jihwan  and
    Lee, Jae-Gil",
    booktitle = "Proceedings of the 41st International Conference on Machine Learning",
    year = "2024",
    address = "Vienna, Austria",
    publisher = "JMLR.org",
    url = "https://dl.acm.org/doi/10.5555/3692070.3693057",
    articleno = "987",
    numpages = "16"
}

\clearpage
\appendix
\section*{Appendix}

\section{Dataset Details}
\label{sec:dataset_details}
Table~\ref{tab:datasets} summarizes the datasets used across question answering and summarization tasks, divided into in-domain and out-of-domain splits. For summarization, the numbers of training, validation, and test samples are capped at 40K, 10K, and 3K per task, respectively.

\begin{table}[H]
\centering
\scriptsize
\begin{tabular}{lll}
\toprule
\textbf{Split} & \textbf{Dataset} & \textbf{Abbrev.} \\
\midrule

\multicolumn{3}{c}{\textbf{Question Answering}} \\
\midrule

In & SQuAD~\citep{rajpurkar-etal-2016-squad} & SQ \\
In & NewsQA~\citep{trischler-etal-2017-newsqa} & News \\
In & TriviaQA~\citep{joshi-etal-2017-triviaqa} & Tri \\
In & SearchQA~\citep{dunn-etal-2017-searchqa} & Srch \\
In & HotpotQA~\citep{yang-etal-2018-hotpotqa} & HP \\
In & Natural Questions~\citep{kwiatkowski-etal-2019-natural} & NQ \\

Out & BioASQ~\citep{tsatsaronis-etal-2012-bioasq} & BSQ \\
Out & DROP~\citep{dua-etal-2019-drop} & DP \\
Out & DuoRC~\citep{saha-etal-2018-duorc} & DRC \\
Out & RACE~\citep{lai-etal-2017-race} & RC \\
Out & RelationExtraction~\citep{levy-etal-2017-zero} & RE \\
Out & TextbookQA~\citep{kembhavi-etal-2017-are} & TB \\

\midrule

\multicolumn{3}{c}{\textbf{Summarization}} \\
\midrule

In & WikiLingua~\citep{ladhak-etal-2020-wikilingua} & Wiki \\
In & Reddit TIFU~\citep{volske-etal-2017-tl} & Red \\
In & XSum~\citep{narayan-etal-2018-dont} & XS \\
In & CNN/DailyMail~\citep{hermann-etal-2015-teaching} & CNN \\
In & Gigaword~\citep{graff-etal-2003-english} & Giga \\
In & MediaSum~\citep{zhu-etal-2021-mediasum} & Media \\

Out & DialogSum~\citep{chen-etal-2021-dialogsum} & Dialog \\
Out & SciTLDR~\citep{cachola-etal-2020-tldr} & Sci \\
Out & AESLC~\citep{zhang-tetreault-2019-email} & AESLC \\
Out & WikiHow~\citep{koupaee-etal-2018-wikihow} & WH \\
Out & Multi-News~\citep{fabbri-etal-2019-multi} & MN \\

\bottomrule
\end{tabular}
\caption{Datasets used in our experiments.}
\label{tab:datasets}
\end{table}

\section{Input Templates}

All inputs are formatted using a unified conversational template based on the Llama-3.2-1B-Instruct chat format. For question answering tasks, inputs are formatted as:

\begin{quote}
\footnotesize
\ttfamily
<|start\_header\_id|>user<|end\_header\_id|> \\
Context: \{context\} \\
Question: \{question\} \\
<|eot\_id|> \\
<|start\_header\_id|>assistant \\
<|end\_header\_id|> \\
Answer:
\end{quote}

For summarization tasks, the following template is used:

\begin{quote}
\footnotesize
\ttfamily
<|start\_header\_id|>user<|end\_header\_id|> \\
Document: \{document\} \\
<|eot\_id|> \\
<|start\_header\_id|>assistant \\
<|end\_header\_id|> \\
Summary:
\end{quote}

During training, the target sequence is appended after the assistant prefix, while during inference it is excluded.

\section{Parameter Initialization}
\label{sec:initial}
To improve optimization stability, soft prompts are initialized from natural language templates rather than random vectors. The base prompt $P_B$ and projection matrix $W$ are initialized using generic task-independent instructions to reduce task-specific bias:
\begin{quote}
\small
\texttt{[
"Process input and infer task.",\\
"Select relevant information.",\\
"Model relationships in the input.",\\
"Maintain consistency with the input.",\\
"Generate an appropriate response."
]}
\end{quote}

Given the averaged template embedding matrix $E \in \mathbb{R}^{l \times d}$, SVD is applied:
\[
E = U \Sigma V^\top.
\]

The low-rank initialization is constructed using the truncated SVD components, where $r$ denotes the target low-rank dimension:
\[
P_B = U_r \Sigma_r^{1/2}, \qquad W = \Sigma_r^{1/2} V_r^\top.
\]

Task prompts are initialized from task-specific template embeddings projected into the shared low-rank space. For each task $\tau$, the task-specific embedding matrix $E_\tau \in \mathbb{R}^{l \times d}$ is extracted from the frozen backbone and projected via $W$:
\[
P_T^{(\tau)} = E_\tau W^\top.
\]

Prior to Phase 2, the task-specific embedding matrix is mean-pooled to obtain $\bar{E}_\tau \in \mathbb{R}^{d}$, and the task key is initialized as:
\[
k_T^{(\tau)} = \bar{E}_\tau W^\top.
\]

\section{Implementation Details}

\subsection{Experiment Settings}

Table~\ref{tab:generation_settings} summarizes the training and generation settings used across question answering and summarization experiments.

\begin{table}[H]
\centering
\scriptsize
\begin{tabular}{lcc}
\toprule
\textbf{Setting} & \textbf{MRQA} & \textbf{Summarization} \\
\midrule

GPU & \multicolumn{2}{c}{1× RTX PRO 6000 Blackwell} \\
Batch size & \multicolumn{2}{c}{16} \\
Gradient accumulation steps & \multicolumn{2}{c}{16} \\

\midrule

Max sequence length & 512 & 768 \\
Max new tokens & 100 & 128 \\

\midrule

Training steps & \multicolumn{2}{c}{1000} \\
Warm-up steps & \multicolumn{2}{c}{100} \\
Evaluation interval & \multicolumn{2}{c}{Every 100 steps} \\

\midrule

Number of beams & \multicolumn{2}{c}{1} \\
Do sample & \multicolumn{2}{c}{False} \\

\bottomrule
\end{tabular}
\caption{
Training and generation settings used for MRQA and Summarization experiments.
For \algname{}, evaluation is performed only during Phase 2.
}
\label{tab:generation_settings}
\end{table}

\subsection{Baseline Hyperparameter Settings}
\label{sec:baseline_hyperparameters}
Table~\ref{tab:baseline_details} summarizes the main hyperparameter settings used in our experiments. For fair comparison, learning rates followed common settings for each method family, prompt lengths were fixed across methods, and low-rank dimensions were adjusted for comparable parameter budgets. For MoE-based baselines, the number of prompts matched the number of group prompts in \algname{}.

\vspace{-0.3em}

\begin{table*}[!t]
\centering
\scriptsize
\renewcommand{\arraystretch}{1.0}

\resizebox{\textwidth}{!}{
\begin{tabular}{ll}
\toprule
\textbf{Method} & \textbf{Hyperparameter Settings} \\
\midrule

FT
& learning rate=3e-5 \\

LoRA
& learning rate=3e-4; $r=1$; $\alpha=16$; target modules=\{q\_proj, v\_proj\} \\

HydraLoRA
& learning rate=3e-4; $r=1$; $\alpha=16$; expert branches=2; target modules=\{q\_proj, v\_proj\} \\

PT
& learning rate=3e-3; prompt length=40; initialization=template \\

DPT
& learning rate=3e-3; prompt length=40; low-rank dim=42; initialization=template \\

SMoP
& learning rate=3e-3; total prompts=40; prompts=2; initialization=template \\

PT-MoE
& learning rate=3e-3; prompt length=40; prompts=2; low-rank dim=39; initialization=template \\

HiPro-adapted
& learning rate=3e-3; Stage A/B training steps=200/800; prompt length=40; low-rank dim=36\\

\algname{}
& learning rate=3e-3; prompt length=40; base/group/task prompts=1/2/6; low-rank dim=36; \\
& key coefficient=0.1; ARI threshold=0.9; check interval=10; stability window=5; stability patience=2 \\

\bottomrule
\end{tabular}
}

\caption{
Main hyperparameter settings used for baseline methods and \algname{}.
}
\label{tab:baseline_details}
\end{table*}

\begin{table*}[!t]
\centering
\scriptsize
\setlength{\tabcolsep}{1.2pt}
\renewcommand{\arraystretch}{1.0}
\vspace{-0.5em}

\resizebox{\textwidth}{!}{
\begin{tabular}{
l c
c c c c c c >{\columncolor{gray!8}}c
c c c c c c >{\columncolor{gray!8}}c
}
\toprule
\multirow{2}{*}{\textbf{Method}}
& \multirow{2}{*}{\makecell{\textbf{\# of} \\ \textbf{Params}}}
& \multicolumn{7}{c}{\textbf{In-domain}}
& \multicolumn{7}{c}{\textbf{Out-of-domain}} \\
\cmidrule(lr){3-9}
\cmidrule(lr){10-16}
&
& SQ & News & Tri & Srch & HP & NQ & Avg
& BSQ & DP & DRC & RC & RE & TB & Avg \\
\midrule

FT
& 1.2B
& $79.3\std{1.3}$
& $44.9\std{0.6}$
& $65.6\std{1.8}$
& $76.0\std{0.6}$
& $59.9\std{0.7}$
& $60.9\std{0.4}$
& $64.4\std{0.4}$
& $58.1\std{4.0}$
& $40.0\std{1.3}$
& $40.1\std{2.2}$
& $34.6\std{0.8}$
& $74.8\std{1.8}$
& $48.7\std{2.3}$
& $49.4\std{1.7}$
\\

\midrule

LoRA
& 106K
& $79.0\std{0.6}$
& $43.0\std{0.7}$
& $69.2\std{0.5}$
& $72.6\std{1.2}$
& $58.3\std{0.4}$
& $60.2\std{0.2}$
& $63.7\std{0.4}$
& $61.7\std{0.1}$
& $38.6\std{1.3}$
& $40.3\std{0.3}$
& $39.3\std{1.1}$
& $73.2\std{1.0}$
& $51.5\std{1.3}$
& $50.8\std{0.7}$
\\

HydraLoRA
& 278K
& $79.8\std{0.9}$
& $42.5\std{1.0}$
& $69.3\std{0.6}$
& $73.1\std{1.8}$
& $58.9\std{0.5}$
& $60.8\std{0.2}$
& $64.0\std{0.5}$
& $60.8\std{2.4}$
& $39.9\std{1.8}$
& $40.4\std{0.9}$
& $39.2\std{1.3}$
& $73.5\std{0.6}$
& $49.2\std{2.2}$
& $50.5\std{0.7}$
\\

\midrule

PT
& 81K
& $78.6\std{1.0}$
& $39.6\std{2.0}$
& $62.5\std{6.9}$
& $61.3\std{5.2}$
& $56.2\std{0.6}$
& $59.3\std{0.8}$
& $59.6\std{2.0}$
& $59.5\std{2.8}$
& $39.6\std{1.5}$
& $37.5\std{0.2}$
& \underline{$38.5$}$\std{0.2}$
& \underline{$74.9$}$\std{0.8}$
& $46.3\std{3.8}$
& $49.4\std{0.2}$
\\

DPT
& 87K
& $78.4\std{0.3}$
& $38.3\std{4.0}$
& \underline{$64.6$}$\std{3.7}$
& $68.7\std{3.9}$
& \underline{$57.0$}$\std{1.0}$
& $59.4\std{1.1}$
& $61.1\std{2.3}$
& \underline{$61.0$}$\std{4.0}$
& \underline{$42.0$}$\std{0.6}$
& $39.3\std{0.3}$
& $37.5\std{0.9}$
& $73.4\std{1.5}$
& \underline{$49.9$}$\std{0.5}$
& \underline{$50.5$}$\std{1.2}$
\\

SMoP
& 86K
& \underline{$79.4$}$\std{1.2}$
& $38.5\std{3.6}$
& $63.4\std{2.7}$
& \underline{$71.4$}$\std{1.3}$
& $56.9\std{0.8}$
& \underline{$59.9$}$\std{1.2}$
& \underline{$61.6$}$\std{0.8}$
& $61.0\std{3.2}$
& $41.8\std{1.0}$
& $38.1\std{2.1}$
& $37.6\std{2.8}$
& $74.4\std{1.0}$
& $47.3\std{2.3}$
& $50.0\std{1.2}$
\\

PT-MoE
& 87K
& $78.5\std{0.9}$
& \underline{$40.6$}$\std{0.8}$
& $63.2\std{4.3}$
& $66.5\std{3.1}$
& $57.0\std{0.2}$
& $58.5\std{1.2}$
& $60.7\std{0.6}$
& $59.9\std{1.7}$
& $36.5\std{0.8}$
& $\mathbf{40.5}\std{0.9}$
& $\mathbf{38.2}\std{0.4}$
& $74.0\std{1.0}$
& $48.2\std{0.7}$
& $49.6\std{0.3}$
\\

\rowcolor{gray!10}
\textbf{\algname{}}
& 87K
& $\mathbf{79.8}\std{1.1}$
& $\mathbf{42.8}\std{1.0}$
& $\mathbf{67.9}\std{4.2}$
& $\mathbf{73.6}\std{0.8}$
& $\mathbf{59.2}\std{0.2}$
& $\mathbf{61.9}\std{0.3}$
& $\mathbf{64.2}\std{0.5}$
& $\mathbf{64.0}\std{2.0}$
& $\mathbf{44.6}\std{0.2}$
& \underline{$39.6$}$\std{0.7}$
& $38.0\std{1.4}$
& $\mathbf{74.6}\std{2.2}$
& $\mathbf{52.1}\std{1.2}$
& $\mathbf{52.1}\std{0.7}$
\\

\bottomrule
\end{tabular}
}

\vspace{-0.5em}
\caption{
MRQA EM comparison on in-domain and out-of-domain datasets with mean and standard deviation.
}
\label{tab:em_results_std}

\end{table*}

\begin{table*}[!t]
\centering
\scriptsize
\setlength{\tabcolsep}{6.7pt}
\renewcommand{\arraystretch}{1.0}
\vspace{-0.5em}

\resizebox{\textwidth}{!}{
\begin{tabular}{
l c
c c c c c c >{\columncolor{gray!8}}c
c c c c c c >{\columncolor{gray!8}}c
}
\toprule
\multirow{2}{*}{\textbf{Method}}
& \multirow{2}{*}{\makecell{\textbf{\# of} \\ \textbf{Params}}}
& \multicolumn{7}{c}{\textbf{In-domain}}
& \multicolumn{7}{c}{\textbf{Out-of-domain}} \\
\cmidrule(lr){3-9}
\cmidrule(lr){10-16}
&
& SQ & News & Tri & Srch & HP & NQ & Avg
& BSQ & DP & DRC & RC & RE & TB & Avg \\
\midrule

SMoP
& 172K
& \underline{94.8}
& \textbf{68.6}
& 85.4
& \underline{88.8}
& \underline{82.3}
& 80.0
& \underline{83.3}
& \textbf{82.1}
& \underline{71.3}
& \underline{57.7}
& \textbf{65.8}
& 87.9
& 72.2
& \underline{72.8}
\\

PT-MoE
& 171K
& 94.3
& 68.0
& \underline{85.7}
& 88.3
& 80.8
& \underline{80.4}
& 82.9
& \underline{81.8}
& 67.9
& 56.0
& 63.0
& \underline{88.2}
& \underline{74.3}
& 71.9
\\

\rowcolor{gray!12}
\textbf{\algname{}}
& 157K
& \textbf{95.4}
& \underline{68.0}
& \textbf{85.8}
& \textbf{89.8}
& \textbf{82.6}
& \textbf{82.3}
& \textbf{84.2}
& 81.4
& \textbf{74.7}
& \textbf{58.2}
& \underline{65.2}
& \textbf{88.3}
& \textbf{74.6}
& \textbf{73.7}
\\

\bottomrule
\end{tabular}
}

\vspace{-0.5em}
\caption{
MRQA F1 comparison on in-domain and out-of-domain datasets with Llama-3.1-8B-Instruct.
}
\label{tab:8B_mrqa_results}

\end{table*}

\begin{table*}[!t]
\centering
\scriptsize
\setlength{\tabcolsep}{6.7pt}
\renewcommand{\arraystretch}{1.0}
\vspace{-0.5em}

\resizebox{\textwidth}{!}{
\begin{tabular}{
l c
c c c c c c >{\columncolor{gray!8}}c
c c c c c >{\columncolor{gray!8}}c
}
\toprule
\multirow{2}{*}{\textbf{Method}}
& \multirow{2}{*}{\makecell{\textbf{\# of} \\ \textbf{Params}}}
& \multicolumn{7}{c}{\textbf{In-domain}}
& \multicolumn{6}{c}{\textbf{Out-of-domain}} \\
\cmidrule(lr){3-9}
\cmidrule(lr){10-15}
&
& Wiki & Red & XS & CNN & Giga & Media & Avg
& Dialog & Sci & AESLC & WH & MN & Avg \\
\midrule

SMoP
& 172K
& \underline{36.3}
& \underline{23.5}
& \underline{33.1}
& 23.2
& \underline{41.8}
& 16.9
& \underline{29.1}
& \underline{18.7}
& 17.9
& \underline{17.8}
& \textbf{25.6}
& 10.3
& 18.1
\\

PT-MoE
& 171K
& 35.5
& 23.0
& 32.7
& \underline{23.5}
& 41.4
& \underline{17.2}
& 28.9
& 18.1
& \underline{19.7}
& \textbf{17.9}
& 24.8
& \underline{10.4}
& \underline{18.2}
\\

\rowcolor{gray!12}
\textbf{\algname{}}
& 159K
& \textbf{37.4}
& \textbf{24.1}
& \textbf{34.0}
& \textbf{26.9}
& \textbf{42.4}
& \textbf{18.7}
& \textbf{30.6}
& \textbf{19.8}
& \textbf{20.6}
& 15.5
& \underline{25.0}
& \textbf{13.4}
& \textbf{18.9}
\\

\bottomrule
\end{tabular}
}

\vspace{-0.5em}
\caption{
Summarization ROUGE-L comparison on in-domain and
out-of-domain datasets with Llama-3.1-8B-Instruct.
}
\label{tab:8B_summarization_results}

\end{table*}

\begin{table}[t]
\centering
\scriptsize
\setlength{\tabcolsep}{6pt}
\renewcommand{\arraystretch}{1.2}
\begin{tabular}{lccc}
\toprule
\textbf{Method} & \textbf{Prompt Length} & \textbf{Relative FLOPs} & \textbf{Latency (ms)} \\
\midrule
PT      & 40    & 1.1624$\times$ & 38.17 \\
DPT     & 40    & 1.1624$\times$ & 44.78 \\
SMoP    & 20    & 1.0797$\times$ & 40.98 \\
PT-MoE  & 40    & 1.1624$\times$ & 38.43 \\
\rowcolor{gray!10}
\textbf{\algname{}} & 40 & 1.1624$\times$ & 38.38 \\
\bottomrule
\end{tabular}
\caption{Inference efficiency comparison with prompt-tuning baselines.}
\label{table:efficiency}
\end{table}

\section{Additional Experiments}

\subsection{MRQA EM Results}
\label{sec:mrqa_em_results}

Table~\ref{tab:em_results_std} presents detailed MRQA exact match (EM) results across in- and out-of-domain datasets.

\subsection{Additional Backbone Experiments}
Tables~\ref{tab:8B_mrqa_results} and~\ref{tab:8B_summarization_results} present additional results using Llama-3.1-8B-Instruct, showing that \algname{} achieves higher average performance on both benchmarks while using fewer parameters.

\subsection{Efficiency Analysis}
We additionally compare relative FLOPs and inference latency to evaluate the efficiency of \algname{}, as shown in Table~\ref{table:efficiency}. Overall, \algname{} achieves inference efficiency comparable to existing prompt-tuning baselines.

\section{Detailed Ablation Results}
\label{appendix:ablation}

\subsection{Per-Task Ablation Results}

Tables~\ref{tab:appendix_mrqa_ablation} and~\ref{tab:appendix_summary_ablation} present full ablation results on MRQA and summarization benchmarks.

\begin{table*}[!t]
\centering
\scriptsize
\setlength{\tabcolsep}{1.7pt}
\renewcommand{\arraystretch}{1.0}
\vspace{-0.5em}

\begin{tabular}{l ccccccc c ccccccc}
\toprule
\multirow{2}{*}{\textbf{Method}}
& \multicolumn{7}{c}{\textbf{In-domain}}
& \multicolumn{7}{c}{\textbf{Out-of-domain}} \\
\cmidrule(lr){2-8}
\cmidrule(lr){9-15}
&
SQ & News & Tri & Srch & HP & NQ & Avg
& BSQ & DP & DRC & RC & RE & TB & Avg \\
\midrule

\rowcolor{gray!10}
\algname{}
& $87.7\std{0.8}$
& $59.2\std{0.9}$
& $74.9\std{1.7}$
& $80.5\std{0.5}$
& $75.2\std{0.6}$
& $74.9\std{0.5}$
& $75.4\std{0.3}$
& $78.3\std{0.9}$
& $53.9\std{0.2}$
& $48.9\std{0.4}$
& $50.1\std{0.8}$
& $86.0\std{1.0}$
& $59.8\std{0.9}$
& $62.8\std{0.3}$
\\

\midrule

\textit{w/o Residual Decomp.}
& $87.4\std{0.3}$
& $58.7\std{2.0}$
& $74.0\std{1.4}$
& $80.2\std{0.2}$
& $74.8\std{0.7}$
& $74.4\std{0.5}$
& $74.9\std{0.5}$
& $76.5\std{1.4}$
& $51.8\std{0.6}$
& $48.0\std{1.1}$
& $49.7\std{1.4}$
& $85.9\std{0.6}$
& $59.2\std{0.7}$
& $61.8\std{0.2}$
\\

\midrule

\multicolumn{15}{c}{\textit{w/o Hierarchical Structure}} \\
\midrule

Base-level only
& $87.0\std{0.2}$
& $57.0\std{1.4}$
& $73.8\std{0.8}$
& $77.3\std{1.3}$
& $73.2\std{0.3}$
& $72.9\std{0.7}$
& $73.6\std{0.7}$
& $78.0\std{0.4}$
& $50.9\std{1.1}$
& $48.7\std{0.8}$
& $50.2\std{0.8}$
& $86.3\std{0.3}$
& $58.7\std{1.3}$
& $62.1\std{0.2}$
\\

Group-level only
& $86.7\std{0.9}$
& $59.1\std{0.6}$
& $72.7\std{1.3}$
& $77.7\std{1.0}$
& $73.2\std{0.6}$
& $72.9\std{0.3}$
& $73.7\std{0.7}$
& $77.1\std{0.5}$
& $49.5\std{0.5}$
& $49.1\std{0.8}$
& $49.7\std{1.2}$
& $85.7\std{0.6}$
& $55.8\std{2.7}$
& $61.1\std{0.8}$
\\

Task-level only
& $88.2\std{0.4}$
& $57.5\std{1.3}$
& $73.1\std{1.9}$
& $78.8\std{1.4}$
& $74.5\std{0.5}$
& $74.5\std{0.3}$
& $74.4\std{0.6}$
& $76.9\std{0.1}$
& $53.3\std{0.4}$
& $46.6\std{1.0}$
& $47.9\std{0.8}$
& $86.5\std{0.2}$
& $58.7\std{1.4}$
& $61.6\std{0.4}$
\\

\midrule

\multicolumn{15}{c}{\textit{w/o Vertical Mixture-of-Experts}} \\
\midrule

Base-key only
& $85.2\std{0.5}$
& $56.7\std{0.3}$
& $76.8\std{0.3}$
& $67.5\std{1.0}$
& $69.9\std{1.5}$
& $70.4\std{0.2}$
& $71.1\std{0.1}$
& $80.3\std{0.8}$
& $53.3\std{0.3}$
& $48.6\std{0.1}$
& $50.5\std{0.4}$
& $87.2\std{0.2}$
& $58.4\std{0.2}$
& $63.1\std{0.2}$
\\

Group-key only
& $86.4\std{0.3}$
& $59.6\std{0.3}$
& $78.9\std{0.8}$
& $77.3\std{2.7}$
& $72.5\std{0.3}$
& $72.2\std{0.1}$
& $74.5\std{0.5}$
& $78.4\std{0.3}$
& $52.5\std{0.5}$
& $49.3\std{0.3}$
& $51.5\std{0.0}$
& $86.7\std{0.6}$
& $57.1\std{0.1}$
& $62.6\std{0.1}$
\\

Task-key only
& $88.3\std{0.6}$
& $58.7\std{0.2}$
& $72.9\std{2.3}$
& $80.7\std{0.4}$
& $75.4\std{0.5}$
& $74.7\std{0.1}$
& $75.1\std{0.2}$
& $78.5\std{0.1}$
& $53.2\std{0.1}$
& $48.3\std{0.1}$
& $49.9\std{0.2}$
& $86.9\std{0.1}$
& $59.4\std{0.2}$
& $62.7\std{0.1}$
\\

\bottomrule
\end{tabular}

\vspace{0.3em}
\caption{
Detailed ablation study of \algname{} on in-domain and out-of-domain MRQA datasets.
}
\label{tab:appendix_mrqa_ablation}
\end{table*}

\begin{table*}[!t]
\centering
\scriptsize
\setlength{\tabcolsep}{2.7pt}
\renewcommand{\arraystretch}{1.0}
\vspace{-0.5em}

\begin{tabular}{l ccccccc cccccc}
\toprule
\multirow{2}{*}{\textbf{Method}}
& \multicolumn{7}{c}{\textbf{In-domain}}
& \multicolumn{6}{c}{\textbf{Out-of-domain}} \\
\cmidrule(lr){2-8}
\cmidrule(lr){9-14}
&
Wiki & Red & XS & CNN & Giga & Media & Avg
& Dialog & Sci & AESLC & WH & MN & Avg \\
\midrule

\rowcolor{gray!10}
\textbf{\algname{}}
& $30.5\std{0.3}$
& $20.6\std{0.7}$
& $25.9\std{0.2}$
& $24.0\std{0.5}$
& $38.8\std{0.1}$
& $16.7\std{2.5}$
& $26.1\std{0.6}$
& $17.5\std{1.6}$
& $18.7\std{2.0}$
& $14.4\std{0.1}$
& $20.9\std{1.0}$
& $12.3\std{0.6}$
& $16.8\std{0.3}$
\\

\midrule

\textit{w/o Residual Decomp.}
& $30.7\std{0.1}$
& $20.9\std{0.2}$
& $26.2\std{0.4}$
& $24.4\std{0.6}$
& $38.6\std{0.4}$
& $15.5\std{0.8}$
& $26.0\std{0.3}$
& $16.1\std{0.9}$
& $15.1\std{0.4}$
& $10.3\std{0.1}$
& $17.0\std{1.4}$
& $13.3\std{0.8}$
& $14.4\std{0.4}$
\\

\midrule

\multicolumn{14}{c}{\textit{w/o Hierarchical Structure}} \\
\midrule

Base-level only
& $29.3\std{0.8}$
& $20.3\std{0.5}$
& $25.1\std{0.2}$
& $21.7\std{0.6}$
& $38.2\std{0.8}$
& $17.6\std{0.4}$
& $25.4\std{0.1}$
& $18.9\std{0.1}$
& $18.9\std{1.3}$
& $13.9\std{0.5}$
& $19.1\std{0.1}$
& $9.2\std{1.1}$
& $16.0\std{0.4}$
\\

Group-level only
& $29.7\std{0.1}$
& $20.4\std{0.1}$
& $24.8\std{0.6}$
& $23.2\std{0.6}$
& $38.2\std{0.6}$
& $16.9\std{0.1}$
& $25.5\std{0.1}$
& $18.8\std{0.3}$
& $17.7\std{1.8}$
& $11.6\std{0.8}$
& $19.6\std{3.0}$
& $8.3\std{1.1}$
& $15.2\std{1.0}$
\\

Task-level only
& $31.1\std{0.3}$
& $20.8\std{0.4}$
& $26.1\std{0.2}$
& $24.7\std{1.2}$
& $39.2\std{0.1}$
& $14.8\std{1.0}$
& $26.1\std{0.2}$
& $17.0\std{0.5}$
& $15.4\std{0.2}$
& $10.1\std{1.0}$
& $16.0\std{1.7}$
& $14.2\std{0.1}$
& $14.5\std{0.6}$
\\

\midrule

\multicolumn{14}{c}{\textit{w/o Vertical MoE}} \\
\midrule

Base-key only
& $17.8\std{0.6}$
& $15.9\std{0.4}$
& $21.0\std{1.1}$
& $21.5\std{0.4}$
& $28.1\std{1.3}$
& $18.5\std{0.4}$
& $20.5\std{0.4}$
& $20.0\std{0.1}$
& $18.8\std{0.6}$
& $14.5\std{0.5}$
& $13.0\std{0.4}$
& $8.9\std{0.4}$
& $15.0\std{0.1}$
\\

Group-key only
& $18.4\std{2.1}$
& $17.1\std{1.8}$
& $23.0\std{0.4}$
& $19.7\std{1.5}$
& $27.6\std{1.0}$
& $19.2\std{1.4}$
& $20.9\std{0.8}$
& $19.1\std{0.5}$
& $19.6\std{2.1}$
& $14.1\std{0.6}$
& $13.6\std{3.0}$
& $7.7\std{1.1}$
& $14.8\std{0.1}$
\\

Task-key only
& $31.1\std{0.4}$
& $20.7\std{0.6}$
& $26.1\std{0.2}$
& $24.2\std{0.6}$
& $39.0\std{0.1}$
& $15.5\std{0.9}$
& $26.1\std{0.3}$
& $16.3\std{1.3}$
& $15.3\std{0.9}$
& $14.0\std{1.9}$
& $21.0\std{1.3}$
& $13.3\std{0.4}$
& $16.0\std{0.2}$
\\

\bottomrule
\end{tabular}

\vspace{0.3em}
\caption{
Detailed ablation study of \algname{} on in-domain and out-of-domain summarization datasets.
}
\label{tab:appendix_summary_ablation}
\end{table*}

\subsection{Effect of Residual Decomposition}

\begin{figure}[t]
    \centering
    \includegraphics[width=\columnwidth]{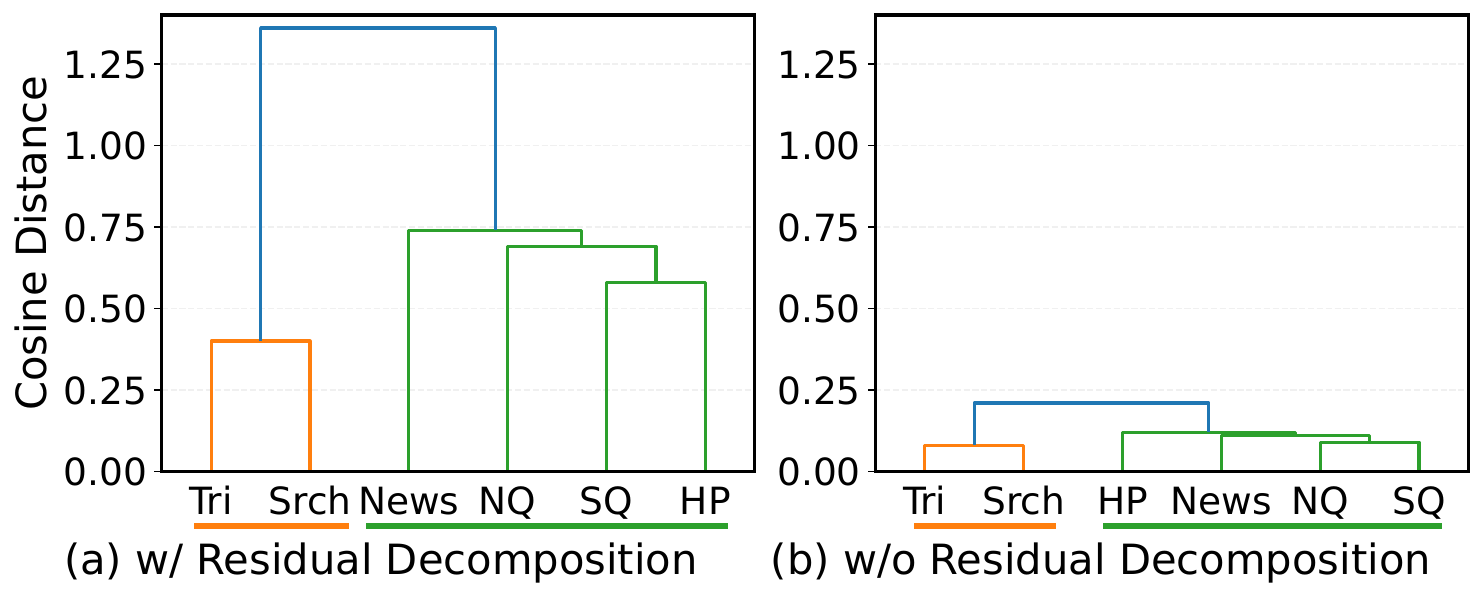}
    \caption{
    Hierarchical clustering structures with and without residual decomposition on MRQA tasks. Residual decomposition yields clearer task separation while preserving overall grouping structure.
    }
    \label{fig:residual_grouping}
\end{figure}

Figure~\ref{fig:residual_grouping} compares the hierarchical clustering structures learned with and without residual decomposition on MRQA tasks. Both variants produce similar high-level grouping structures, indicating that the underlying semantic relationships between tasks are preserved. In contrast, the variant with residual decomposition exhibits clearer separation between task groups, suggesting that residual decomposition more effectively removes shared prompt components while preserving task-specific structure.

\subsection{Effect of Grouping Strategy}
\label{sec:grouping_strategy}

We further investigate whether the improvement comes from introducing
a group level or from the proposed stability-driven grouping strategy.
Table~\ref{tab:grouping_strategy} compares HiVe with a two-level
Base+Task variant and a Step-0 static variant, which fixes task groups
using initial data representations before prompt training. Stability-driven grouping performs better than both alternatives, suggesting that grouping tasks based on stabilized task-prompt representations is more effective than either omitting the group level or fixing the grouping before inter-task relationships are learned.

\begin{table}[t]
\centering
\small
\begin{tabular}{lcc}
\toprule
Variant & In-domain & OOD \\
\midrule
Base+Task
& $74.8\std{0.3}$ & $61.9\std{0.4}$ \\
Step-0 static grouping
& $74.4\std{0.5}$ & $61.6\std{0.6}$ \\
\rowcolor{gray!10}
Stability-driven grouping
& $\mathbf{75.4}\std{0.3}$ & $\mathbf{62.8}\std{0.3}$ \\
\bottomrule
\end{tabular}
\caption{Effect of grouping strategy on MRQA. The default variant is shaded in gray.}
\label{tab:grouping_strategy}
\end{table}

\subsection{Ablation of Key Design Choices}
\label{sec:key_design_ablation}

Table~\ref{tab:key_design_ablation} evaluates alternative choices for task clustering, upper-level key construction, and query--key matching on MRQA. HAC and K-means yield comparable performance, suggesting that grouping based on learned task-prompt representations is robust to the clustering algorithm. In contrast, independently learning the Base and Group keys causes most inputs to match the Base key, substantially reducing in-domain performance. Replacing cosine similarity with negative Euclidean distance also degrades both in-domain and OOD results. These findings support centroid-derived upper-level keys and cosine-based matching.

\begin{table}[t]
\centering
\small
\setlength{\tabcolsep}{2pt}
\begin{tabular}{llcc}
\toprule
Component & Variant & In & OOD \\
\midrule
\multirow{2}{*}{Clustering}
& K-means
& $75.1\std{0.3}$ & $62.8\std{0.1}$ \\
& \cellcolor{gray!10}HAC
& \cellcolor{gray!10}$75.4\std{0.3}$
& \cellcolor{gray!10}$62.8\std{0.3}$ \\
\midrule
\multirow{2}{*}{Upper-level keys}
& Independently learned
& $71.7\std{0.7}$ & $62.5\std{0.4}$ \\
& \cellcolor{gray!10}Centroid-derived
& \cellcolor{gray!10}$\mathbf{75.4}\std{0.3}$
& \cellcolor{gray!10}$\mathbf{62.8}\std{0.3}$ \\
\midrule
\multirow{2}{*}{Matching metric}
& Negative Euclidean
& $74.0\std{0.5}$ & $62.3\std{0.2}$ \\
& \cellcolor{gray!10}Cosine
& \cellcolor{gray!10}$\mathbf{75.4}\std{0.3}$
& \cellcolor{gray!10}$\mathbf{62.8}\std{0.3}$ \\
\bottomrule
\end{tabular}
\caption{Ablation of key design choices on MRQA. Default variants are shaded in gray.}
\label{tab:key_design_ablation}
\end{table}

\end{document}